\documentclass[11pt]{article}
\usepackage{style}
 \usepackage{authblk}
\crefname{claim}{claim}{claims}  
\Crefname{claim}{Claim}{Claims} 

\title{On learning functions over biological sequence space: relating Gaussian process priors, regularization, and gauge fixing}
\date{}

\begin{document}

\author[1\footnote{E-mail: \href{mailto:samantha.petti@tufts.edu}{samantha.petti@tufts.edu}}]{Samantha Petti}
\author[2]{Carlos Mart\'i-G\'omez}
\author[2]{Justin B. Kinney}
\author[3]{Juannan Zhou}
\author[2]{David M. McCandlish}

\affil[1]{Department of Mathematics, Tufts University, Medford, MA, 02155. }
\affil[2]{Simons Center for Quantitative Biology, Cold Spring Harbor Laboratory, Cold Spring Harbor, NY, 11724}
\affil[3]{Department of Biology, University of Florida, Gainesville, FL, 32611}

\maketitle

\begin{abstract}
\noindent Mappings from biological sequences (DNA, RNA, protein) to quantitative measures of sequence functionality play an important role in contemporary biology. We are interested in the related tasks of (i) inferring predictive sequence-to-function maps and (ii) decomposing sequence-function maps to elucidate the contributions of individual subsequences. Because each sequence-function map can be written as a weighted sum over subsequences in multiple ways, meaningfully interpreting these weights requires “gauge-fixing,” i.e., defining a unique representation for each map. Recent work has established that most existing gauge-fixed representations arise as the unique solutions to $L_2$-regularized regression in an overparameterized ``weight space'' where the choice of regularizer defines the gauge. Here, we establish the relationship between regularized regression in overparameterized weight space and Gaussian process approaches that operate in ``function space,'' i.e.~the space of all real-valued functions on a finite set of sequences.  We disentangle how weight space regularizers both impose an implicit prior on the learned function and restrict the optimal weights to a particular gauge. We show how to construct regularizers that correspond to arbitrary explicit Gaussian process priors combined with a wide variety of gauges and characterize the implicit function space priors associated with the most common weight space regularizers. Finally, we derive the posterior distribution of a broad class of sequence-to-function statistics, including gauge-fixed weights and multiple systems for expressing higher-order epistatic coefficients. We show that such distributions can be efficiently computed for product-kernel priors using a kernel trick.
\end{abstract}

\section{Introduction}
A fundamental goal of biology is to understand how sequence-level differences in DNA, RNA or protein result in different observable outcomes. This mapping from DNA, RNA or protein sequences to some quantitative measure of sequence functionality, e.g.~the growth rate (fitness) of a microbe or binding affinity of a protein, can be difficult to predict and interpret because combinations of mutations interact in complex ways~\cite{Phillips08,sackton2016genotypic,Starr16,Weinreich18,domingo2019causes,zhou2022higher,johnson2023epistasis}. Recent technological advances have produced datasets that can help us characterize such relationships on unprecedented scales. It is now possible to construct libraries of millions of sequences and simultaneously measure an associated function for each sequence \cite{Fowler14,Kinney19,ba2022barcoded}.  This has given rise to interest in method development for characterizing and interpreting sequence-function relationships \cite{beerenwinkel2007epistasis,beerenwinkel2007analysis,hinkley2011systems,poelwijk2016context,sailer2017detecting,otwinowski2018inferring,zhou2020minimum,poelwijk2019learning,tareen2022mave,zhou2022higher,brookes2022sparsity,faure2024mochi,faure2024extension,park2024simplicity,crona2024walsh,marti2025inference}. When successful, such methods can provide biological insight into the mechanisms that determine the function of the sequence and can be applied to practical problems such as protein engineering~\cite{yang2019machine,freschlin2022machine,notin2024machine}.

Here we focus on two related tasks: (i) inferring sequence-function maps, and (ii) decomposing such maps to elucidate the contribution of individual subsequences (including gapped subsequences whose positions are not necessarily contiguous). An accurate approximation of the sequence-function map is useful in that it can be applied to make predictions about the function of sequences that have not been measured as well as to mitigate the influence of measurement noise. Decomposing a sequence-function map into contributions from individual subsequences, which correspond to particular chemical subunits that are present in the cell, is one important strategy for interpreting sequence-function mappings and is particularly useful for suggesting mechanistic hypotheses that explain the observed sequence-function data.
 However, interpreting a decomposition into contributions of individual subsequences is made complicated by a lack of identifiability.  

To see the simplest way in which this lack of identifiability arises, note that in principle we can define one feature per subsequence and hence learn a weight for each. The predicted function of a sequence can then be obtained by summing the weights corresponding to all of its subsequences. For sequences of length $\ell$ on an $\alpha$ character alphabet, we can view the weights as an $(\alpha +1)^\ell$-dimensional vector in \textit{weight space} where each dimension corresponds to a possible subsequence.
This model is non-identifiable: a particular sequence-function map can be expressed by many different weight space vectors because there are only $\alpha^{\ell}$ possible sequences for which to measure values, leading to $(\alpha +1)^\ell-\alpha ^\ell$ extra degrees of freedom. Thus, meaningfully interpreting the weights requires ``fixing the gauge,''~\cite{posfai2024gauge} that is, imposing additional constraints to ensure that the learned vector of weights lies in a particular gauge, where a gauge is defined as a subset of weight space in which each sequence-function map can be expressed uniquely.  Weights have different interpretations in different gauges, and indeed, strategic choices of gauge can help guide the exploration and interpretation of complex functional landscapes \cite{posfai2024gauge}. Importantly, this need to fix the gauge arises even in the simplest models of sequence-function relationships such as pairwise interaction models~\cite{weigt2009identification, ekeberg2013improved, stein2015inferring}, and essentially occurs for any model whose form respects certain symmetries of the space of possible sequences~\cite{posfai2025symmetry}, meaning that the issue of how to appropriately fix the gauge arises quite generally.

Recent work established that gauges that take the form of linear subspaces arise as the set of unique solutions to $L_2$-regularized regression in weight space \cite{posfai2024gauge}, where the choice of positive-definite regularizer specifies the gauge. However, such regularizers result in two very different types of shrinkage. First, a regularizer designed for a particular gauge imposes a penalty on the component of each weight vector that is not in the gauge, which ultimately shrinks that component to zero and hence enforces that the optimal solution lies in the corresponding gauge space. Second, any positive-definite regularizer on the weights also enforces a pattern of shrinkage on the  $\alpha^\ell$-dimensional vector of estimates produced by the regression procedure, where each dimension corresponds to a sequence and the value of a sequence represents our estimate of some measured function of the sequence. That is, by shrinking the estimated values of the weights, the regularizer also produces shrinkage in ``function space,'' but the geometric features and biological interpretation of this shrinkage remains unclear.

A different approach to modeling sequence-function mappings is given by Gaussian process regression~\cite{Rasmussen06}, which works by directly specifying a Bayesian prior over function space. To formulate Gaussian process regression in \textit{function space}, we consider each sequence-function map as an $\alpha^\ell$-dimensional vector where each dimension corresponds to a sequence and the value of a sequence represents some measured function of the sequence. Then we assume the function space vector of measured values is drawn from a multivariate Gaussian distribution. The covariance of this distribution, which is specified by a kernel matrix, expresses a prior belief about which sequences should have similar measured function values. Gaussian process regression yields a posterior distribution over sequence-function maps. A variety of kernels for Gaussian process regression have been proposed and applied to successfully predict sequence-function maps and quantify the uncertainty of predictions \cite{schweikert_empirical_2008,toussaint_exploiting_2010,romero_navigating_2013,yang_learned_2018,zhou2022higher,amin_biological_2023,zhou2024tbd}.

Here, we provide a treatment of the relationship between gauge fixing, $L_2$-regularized linear regression, and Gaussian process regression for learning sequence-function mappings. These connections arise naturally due to the well-known connection between $L_2$-regularized linear regression and Bayesian linear regression in the case where the prior on the weights is multivariate Gaussian. From this point of view, besides fixing the gauge, $L_2$-regularization corresponds to implicitly imposing a Bayesian prior on weight space. We will show how these implicit priors on weight space can be translated into priors on function space, thus clarifying the relationship between regularized regression in $(\alpha+1)^\ell$-dimensional  weight space and Gaussian process regression in $\alpha^\ell$-dimensional function space.

In addition to relating function space Gaussian processes to the implicit priors induced by weight space regularized regression, we consider the gauge-fixed weights of draws from posterior distributions of function space Gaussian processes. These distributions are fundamentally different than posterior distributions of weights achieved via Bayesian regression with a full rank prior: while it is possible to design Gaussian priors on weights such that the maximum a posteriori (MAP) estimate is guaranteed to be in a certain gauge, the support of this posterior is always the entire weight space. This is a problem because the lack of gauge fixing greatly decreases the biological interpretability of these posterior draws.  Analyzing the gauge-fixed weights corresponding to draws from the posterior distribution of a Gaussian process in function space allows us to express uncertainty in the weights while maintaining the interpretability gained by fixing the gauge. Naively, to compute the posterior distribution of gauge-fixed weights or even simply the gauge-fixed weights of the MAP estimate of a function space Gaussian process, one would need to compute an entire sequence-function map and project it into the gauge space of interest. However, for even moderately sized $\alpha$ and $\ell$, this approach is intractable as function space is $\alpha^\ell$-dimensional. We introduce a kernel trick that allows us to efficiently compute any subset of the gauge-fixed weights corresponding to the MAP estimate or posterior distribution for function space Gaussian processes under prior distributions with site factorizable kernel. These results allow us to explicitly specify the function space prior via an appropriate Gaussian process, analyze the results via interpretable gauge-fixed contributions of sub-sequences, and provide uncertainty estimates on these gauge-fixed parameters, all in a computationally tractable manner. Moreover, our kernel trick is more general; it can be applied to compute the posterior distribution of most other previously proposed representations of real-valued functions over sequence space including background-averaged epistasis coefficients~\cite{faure2024extension} and coefficients of the function in the Fourier basis~\cite{brookes2022sparsity}.

\subsection{Summary of our contributions}
Our contributions can be summarized as follows:
\begin{enumerate}
    \item The choice of regularizer for weight space regression both induces a prior on function space and determines the gauge of the optimal solution. We show that for any Gaussian prior on function space and any linear gauge space, there exists a regularizer that induces the prior and whose optimum is in the linear gauge space (\Cref{thm:alway-can-regularize} of \Cref{sec:main-relationship}). Moreover, we establish that such regularizers can be constructed by taking the sum of two matrices: one that determines the prior on function space and the other that determines the gauge. We construct such matrices for priors and gauges of interest in \Cref{sec:build-it}.
    
    \item Diagonal matrices are a natural choice for regularizers in weight space. However, it is not clear a priori what form of shrinkage these regularizers induce in function space. Treating the diagonal regularizers as corresponding to independent zero-mean Gaussian priors on the values of the individual weights, we derive analytic formulas for the function space priors implicitly induced by these diagonal regularizers, and in doing so demonstrate how the choice of diagonal regularizer controls the rate of correlation decay of these induced priors (\Cref{sec:diag}).
    
    \item One way to analyze a complex sequence-function map is to express it in a particular gauge and interpret the weights of each subsequence.  Given  measurements for a fraction of the possible sequences, how do we infer corresponding gauge-fixed weights and quantify the uncertainty? 
    In \Cref{thm:trans-kernel-trick} of \Cref{sec:kernel-trick-gen},  we establish a kernel trick that allows us to efficiently compute the posterior distribution of a large class of linear transformations of functional values without having to explicitly compute an $\alpha^{\ell}$-dimensional sequence-function map. 
      In \Cref{thm:kernel-trick} of \Cref{sec:kernel-trick}, we show how to apply this result to  gauge-fixed weights.
    Given a training set of size $t$, any set of $j$ subsequences, and a function space prior with a product form, we can compute the distribution over $j$ gauge-fixed weights corresponding to the posterior distribution of the function space Gaussian process. Doing so only requires matrix and vector operations with dimensions at most $j$ and $t$.

\end{enumerate}
\subsection{Related work}
The flexibility of Gaussian processes make them an attractive method for modeling complex sequence-relationships so that now many families of kernels have been considered \cite{schweikert_empirical_2008,toussaint_exploiting_2010,romero_navigating_2013,yang_learned_2018,zhou2022higher,amin_biological_2023}.
Here we focus on isotropic kernels and non-isotropic product kernels in which each feature corresponds to a sequence position, see \cite{zhou2022higher,zhou2024tbd}. Due to the mathematical structure of such kernels and recent advances in GPU acceleration~\cite{gardner2018gpytorch, wang2019exact}, these kernels are tractable for inference with hundreds of thousands of sequences and have been shown to exhibit state-of-the-art predictive performance \cite{zhou2022higher,zhou2024tbd}. 
Our work is a conceptual bridge between these families of function space priors and the theory of gauge-fixing for parameter interpretation explored in \cite{posfai2024gauge}.

Our work also provides a bridge between the inference and analysis of empirical sequence-function relationships and the theoretical literature on ``fitness landscapes'' \cite{manrubia2021genotypes,bank2022epistasis}. In particular, special cases of the function space and weight space priors we consider here are equivalent to several notable models in the fitness landscape literature including the connectedness model \cite{reddy2021global} (see \Cref{sec:kern-defn}) and the NK and GNK models \cite{kauffman1989nk,buzas2013analysis,nowak2015analysis,hwang2018universality,brookes2022sparsity} (see \Cref{sec:diag}). 

\subsection{Applicability beyond biological sequences}
While our results were developed with biological sequences in mind, they apply to regression and Gaussian processes over arbitrary finite discrete product spaces. We treat each position in a biological sequence as a categorical variable whose value indicates the character at the position; our models do not take into account sequence order. Therefore, we can apply our results to learn functions of any finite number of categorical features, including the common case where all features are binary. Our results apply directly if the number of categories for each feature is constant and can be extended if not. Example applications of prediction over spaces of categorical features include predicting patient outcomes based on the presence or absence of risk factors and treatments, predicting the productivity of microbial communities based on species composition, and making predictions based on responses to multiple choice surveys \cite{orme2009multiple,santner2012statistical, diaz2024global}.

\subsection{Outline}
We begin with a preliminaries section to introduce notation (\Cref{subsec:notation}), define gauge-fixing and highlight how the choice of gauge can guide the interpretation of the function (\Cref{subsec:gauges}), and introduce Gaussian processes on sequence space (\Cref{subsec:gp}). In \Cref{sec:main-relationship} we establish the relationship between regularized regression in weight space and Gaussian process regression in function space. In doing so, we derive a general formula for weight space regularizers in terms of the induced function space prior and the gauge of the optimizer. In \Cref{sec:build-it}, we describe how to design regularizers for gauges of interest and various function space kernels that have been shown to perform well in practice. Then, in \Cref{sec:diag} we describe the function space priors implicitly induced by diagonal weight space regularizers. Finally, in \Cref{sec:kernel-trick-gen}, we describe a kernel trick for computing the distribution of a class of linear transformations of functional values   corresponding to draws from the posterior distributions of Gaussian processes, and in \Cref{sec:kernel-trick} we demonstrate how to apply this kernel trick to compute the posterior distribution of gauge-fixed weights. 

\section{Preliminaries}\label{sec:prelim}
\subsection{Notation.}\label{subsec:notation} Let $\mathcal{A}$ be the alphabet of characters with $\alpha = |\mathcal{A}|$, and let $\ell$ be the length of the sequences. Our goal is to learn mappings of the form $f: \mathcal{A}^\ell \to \R$.
We can equivalently consider each such map as a vector in $\R^{\alpha^\ell}$ indexed by sequences $x \in \mathcal{A}^\ell$. We refer to $\R^{\alpha^\ell}$ as \textit{function space}.

Let $\mathcal{S}$ denote the set of possible subsequences,

$$\mcS = \{ (S,s) : S \subseteq [\ell], s \in \mathcal{A}^{|S|}\},$$

\noindent where $[\ell]= \{1,2,\dots \ell\}$, $S$ denotes the set of positions, and $s$ denotes the sequence of length $|S|$ corresponding to the characters present at those positions. For a sequence $x$, let $x[S]$ denote the characters that appear at the positions in $S$. For example, if $x = abcde$, $x[\{2,5\}] = be$. Let $w_{(S,s)}$ be denote the weight for subsequence $(S,s)$. For ease of notation, when $S = \emptyset$ and $s$ is the empty string, we write $w_\emptyset$ to refer to the corresponding coefficient.  Note $|\mcS| = (\alpha +1)^\ell$. A real-valued function on sequence space can be written as the weighted sum of indicator functions on these subsequences,

$$f(x) = \sum_{(S,s) \in \mcS} w_{(S,s)} \delta_{x[S]=s}.$$

\noindent Alternately, we can write this expression as 

$$f = \Phi w$$

\noindent where $f$ is an $\alpha^\ell$-dimensional vector indexed by sequences with $f_x= f(x)$, $w$ is a $(\alpha+1)^\ell$-dimensional vector of weights indexed by the subsequence features, and $\Phi$ is an $\alpha^\ell \times (\alpha+1)^\ell$ matrix indexed by sequences and subsequence features with $\Phi_{x, (S,s)} = \delta_{x[S]=s}$.  We refer to $\R^{(\alpha+1)^\ell}$ as \textit{weight space}. The non-identifiablity of representing a function as a vector in weight space is clear from the dimensionality; $w \in \R^{(\alpha+1)^\ell}$ and $f \in \R^{\alpha^\ell}$, so there are many $w$ such that $f= \Phi w$.

All vectors and matrices we consider are indexed by sequences or subsequences and the order does not matter. We therefore use the notation $M_U$ to restrict a matrix to the rows corresponding to the elements in the set $U$. Let $X$ be a set of training sequences, and let $y$ be corresponding observed measurements. We use $\Phi_X$ to denote the matrix $\Phi$ restricted to the rows corresponding to sequences in the training set and $f_X$ to denote the function vector restricted to the sequences in the training set. When two sets appear as subscripts, e.g. $M_{X,Y}$ we restrict the rows of $M$ to the elements of $X$ and the columns to the elements in $Y$. We use $\sigma_n^2$ to denote the noise variance.

Throughout we let $W$ and $\Lambda$ be $(\alpha+1)^\ell \times (\alpha+1)^\ell $ dimensional matrices indexed by subsequences, i.e.~indexed by the elements of $\mathcal{S}$. We use $W$ when the matrix is a covariance matrix and $\Lambda$ when the matrix is used as a regularizer. Similarly, we let $K$ and $\Delta$ be $\alpha^\ell \times \alpha^\ell$-dimensional matrices indexed by sequences, and use $K$ when the matrix is a covariance matrix and $\Delta$ when it is used as a regularizer. We let $N(\mu, K)$ denote the multivariate Gaussian distribution with mean $\mu$ and covariance $K$.

\subsection{Gauges}\label{subsec:gauges}

Here we review the theory of gauge-fixing for biological sequences established in \cite{posfai2024gauge}. There are many vectors in weight space that give rise to the same real-valued function on sequence space. For example, given a vector $w$ in weight space, we can add any real value $a$ to the weight for the empty set, $w'_{ \emptyset} = w_{ \emptyset}+a$, and subtract $a$ from the weights for the subsequences on the first position, $w'_{(\{1\}, c)}=w_{(\{1\}, c)}-a$ for each character $c \in \mathcal{A}$. The resultant weight space vector $w'$ produces the same function on sequence space as $w$, $\Phi w = \Phi w'$. The space of gauge freedoms describes all such directions in weight space along which moving does not change the corresponding function on sequence space, i.e.~vectors $g$ such that $\Phi(w+g) = \Phi w$. The following definition gives an equivalent characterization. 
\begin{definition}
    The space of gauge freedoms $G$ is the subspace of weight space defined by
    
    $$G = \{ w \in \R^{(\alpha+1)^\ell}: \Phi w = 0\}.$$
    
\end{definition}

\begin{definition}
   A subspace of weight space $\Theta$ is a linear gauge space if it is complementary to the space of gauge freedoms $G$. 
\end{definition}
\noindent Note that for a fixed linear gauge space $\Theta$, every function on sequence space can be represented as precisely one vector in $\Theta$.

\subsubsection{The $\lambda$-$\pi$ family of gauges}\label{subsec:lam-pi-fam}

We focus on the $\lambda$-$\pi$ family of gauges introduced in \cite{posfai2024gauge}. Every gauge $\Theta^{\lambda,\pi}$ in this family is defined by two parameters: $\lambda  \in [0, \infty]$ and a product distribution defined by assigning probabilities to the characters at each position $\pi(x) = \prod_{p \in [\ell]} \pi^p_{x_p}$. The parameter $\lambda$ controls the relative magnitudes of the weights between longer and shorter subsequences. When $\lambda = 0$ we have the ``trivial gauge'' in which the weights for all subsequences of length less than $\ell$ are zero; as $\lambda$ increases more weight is pushed to the shorter subsequences. The distribution $\pi$ controls the relative magnitude of weights for different subsequences as a function of how likely they are under $\pi$.

Formally, $\lambda$-$\pi$ gauges are tensor products of single-position gauges $\Theta^{\lambda, \pi^p}$ of the following form

\begin{equation}\label{eq:building-blocks} V_\lambda = span \left\{ \begin{pmatrix}
\lambda \\
1 \\
\vdots \\
1
\end{pmatrix} \right\},
\quad 
V_\perp^{\pi^p} = \left\{ \begin{pmatrix}
0 \\
v_{c_1} \\
\vdots \\
v_{c_\alpha}
\end{pmatrix}: \sum_{c \in \mathcal{A}} v_c \pi^p_c = 0 \right\}, \quad \Theta^{\lambda, \pi^p} = V_\lambda\oplus V_\perp^{\pi^p},
\end{equation}
\noindent where $\pi^p$ is a distribution over characters at position $p$, $\sum_{c \in \mathcal{A}} \pi^p_c =1$ and $\pi^p_c \geq 0$ for all characters $c \in \mathcal{A}$.
The gauge space $\Theta^{\lambda, \pi}$ is the tensor product

$$\Theta^{\lambda, \pi} = \bigotimes_{p=1}^\ell \Theta^{\lambda, \pi^p}.$$
\noindent The projection matrix into gauge $\Theta^{\lambda, \pi}$ is given by 

\begin{equation}\label{eq:proj} P^{\lambda, \pi}_{(S,s),(T,t)} = 
\prod_{p \in S\cap T}\brac{\delta_{s_p = t_p} -\pi^p_{t_p} \eta} \prod_{p \in S \setminus T} (1- \eta) \prod_{p \in T \setminus S}\pi^p_{t_p} \eta  \prod_{p \not \in S\cup T} \eta
\end{equation}
where $\eta = \lambda/(1+\lambda)$ (see \cite{posfai2024gauge}). Note that projection matrix for $\lambda = \infty$ is well-defined as this simply corresponds to the case $\eta = 1$. 

We now highlight two gauges of interest in the $\lambda$-$\pi$ family; for further discussion of these and other specific gauges in the family see \cite{posfai2024gauge}. Additionally, in \Cref{subsec:marg-prop} we establish a new marginalization property for $\lambda$-$\pi$ gauges and use this to further interpret the role of $\lambda$ and $\pi$.
\begin{enumerate} 
    \item Hierarchical gauge. Hierarchical gauges are obtained by taking $\lambda = \infty$ or equivalently $\eta = 1$ in \Cref{eq:proj}. The zero-sum gauge is the hierarchical gauge with $\pi$ as the uniform distribution. In this gauge, the mean function value for sequences with a particular subsequence $(S,s)$ can be expressed simply in terms of the weights. The mean function value over all sequences is $w_{\emptyset}$, the mean function value over all sequences with $c$ at position $p$ is $w_{\emptyset}+w_{(\{p\},c)}$, the mean function value over all sequences with $c$ at position $p$ and $c'$ at position $p'$ is $w_{\emptyset}+w_{(\{p\},c)}+ w_{(\{p'\},c')}+w_{(\{p, p'\},cc')}$, and so on. The function obtained by summing the weights on subsequences of length up to $k$ is the least-squares approximation of the function with up to $k^\mathrm{th}$ order terms. General hierarchical gauges can be interpreted similarly with $\pi$ as the distribution used to compute the mean and least-squares approximation. 
    \item Wild-type gauge. The wild-type gauge is obtained by taking the limit as $\pi$ approaches the probability distribution that has support only at a fixed ``wild-type'' sequence (see \cite{posfai2024gauge} for details). In the wild-type gauge, weights of subsequences that agree with the wild-type sequence at any position are zero, except for $w_{\emptyset}$ which gives the function value of the wild-type sequence. The weight $w_{(\{p\},c)}$ quantifies the effect of the mutation $c$ at position $p$ on the wild-type background, while higher order weights $w_{(S,s)}$ quantify the epistatic (i.e.~interaction) effect due to the group of mutations $(S,s)$.
\end{enumerate}

\subsubsection{Gauge fixing and regularization}
In regularized regression over weight space, the choice of the positive-definite regularizer $\Lambda$ determines the gauge of the optimal solution.
\begin{definition}\label{defn:wopt} Let $\Lambda$ be a positive-definite matrix, and define 

$$w^{OPT}(\Lambda, \beta) = \argmin_{w \in \R^{(\alpha+1)^\ell}} \| y-\Phi_Xw\|_2^2 + \beta w^T \Lambda w.$$ 
We say a positive-definite matrix $\Lambda$ is a $\Theta$-regularizer if $w^{OPT} \in \Theta$ for all $\beta>0$, sequences $X$, and measurements  $y$.
\end{definition}
\noindent In \Cref{sec:build-it}, we will discuss how to construct $\Theta^{\lambda,\pi}$ regularizers. 
\subsection{Gaussian process regression in sequence space.}\label{subsec:gp}

We now outline Gaussian process regression as applied to sequence space. For a comprehensive introduction to Gaussian processes, see \cite{Rasmussen06}. A Gaussian process is defined by a Gaussian prior distribution on functions $f$ given by a covariance matrix $K$ called a \textit{kernel} and a noise variance parameter $\sigma_n^2$.  
We always assume $K$ defines a proper prior, i.e.~$K$ is positive-definite. 

\begin{definition}\label{defn:fmap} Suppose $y = f_X +\ve$ where $f \sim N(0,K)$ and $\ve \sim N(0, \sigma_n^2 I)$. Let $f^{MAP}(K, \sigma_n^2)$ be the Maximum a Posteriori (MAP) estimate for $f$ under this Gaussian process.
\end{definition}

\noindent It is well known that the posterior distribution of $f$ is given by 

\begin{equation}\label{eq:fs-prior}
    f \sim N(K_{*,X} \brac{K_{X,X} + \sigma_n^2 I}^{-1} y, K_{*,*} - K_{*,X} \brac{K_{X,X} + \sigma_n^2 I}^{-1}K_{X,*})
\end{equation}
where the subscript $X$ restricts $K$ to the rows and/or columns corresponding to sequences in the training set $X$ and the subscript $*$ indicates all rows and/or columns, and that moreover the MAP estimate is given by the mean of this posterior distribution (see \cite{Rasmussen06}).


In regularized regression over weight space, the choice of the positive-definite regularizer $\Lambda$ implicitly induces a prior distribution on function space
in addition to determining the gauge of the optimal solution. We say that a regularizer $\Lambda$ induces the prior defined by $K$ if the optimal weights under the regularizer $\Lambda$ yield  the function consistent with the MAP estimate under $K$.

\begin{definition}
We say a regularizer $\Lambda$ induces the prior $K$ if $f^{MAP}(K, \sigma_n^2)= \Phi w^{OPT}(\Lambda, \sigma_n^2)$  for all $\sigma_n^2>0$.
\end{definition}

\noindent In \Cref{sec:kern-defn} we define two broad families of kernels that we will consider: variance component kernels \cite{zhou2022higher} and product kernels \cite{zhou2024tbd}.

\section{The relationship between regularization, Gaussian process priors, and gauge fixing}\label{sec:main-relationship}

In this section we affirmatively resolve the question: for any linear gauge $\Theta$ and prior $K$ on function space, does there exists a $\Theta$-regularizer $\Lambda$ that induces the prior $K$?

\begin{theorem}\label{thm:alway-can-regularize}
    For any linear gauge $\Theta$ and prior $K$ on function space, there exists a $\Theta$-regularizer that induces the prior $K$.  The matrix $\Lambda =\Phi^T K^{-1} \Phi  + B^TB$ where $B$ is a matrix with nullspace $\Theta$ is one such regularizer.
\end{theorem}

The theorem provides a recipe for building a $\Theta$-regularizer that induces a function space prior $K$. Such a regularizer can be written as the sum of two matrices: one that determines the induced prior and one that determines the gauge.  In \Cref{sec:gauges}, we derive a simple matrix of the form $B^TB$ for the $\lambda$-$\pi$ gauges, then in \Cref{sec:priors-in-reg} we compute  $\Phi^T K^{-1} \Phi$ for several useful classes of priors. We can also phrase \Cref{thm:alway-can-regularize} in terms of a Bayesian regression in weight space rather than Gaussian process regression in function space, see \Cref{sec:br-form}.

In \Cref{subsec:gen-cond-for-reg} we 
establish a new general condition for $\Theta$-regularizers that we will apply in the proof of \Cref{thm:alway-can-regularize}.  Then in \Cref{subsec:equiv}  we establish the equivalence of MAP estimates and optimal solutions of regularized regression across function and weight space. In \Cref{sec:br-form} we prove \Cref{thm:alway-can-regularize} and an analogous statement for Bayesian regression. 

 \subsection{Matrices that act as $\Theta$-regularizers}\label{subsec:gen-cond-for-reg}

Posfai et~al.~establish a sufficient condition for a positive-definite matrix to act as a $\Theta$-regularizer (\Cref{defn:wopt}) \cite {posfai2024gauge}.

\begin{definition}
    Let $V_1$ and $V_2$ be complementary subspaces of an $m$-dimensional vector space $V$. An $m \times m$ positive-definite matrix $\Lambda$ orthgonalizes $V_1$ and $V_2$ if for all $v_1 \in V_1$ and $v_2 \in V_2$, 
    
    $$v_1^T \Lambda v_2 =0.$$
    If $\Lambda$ orthgonalizes $V_1$ and $V_2$, we say that $V_1$ and $V_2$ are $\Lambda$-orthogonal. 
\end{definition}

\noindent The following lemma is a special case of Claim 6 of Posfai et al.~\cite{posfai2024gauge}.
\begin{lemma}\label{lem:orthogonalizes-implies-opt-in-gauge} Let $\Theta$ be a gauge space, and let $G$ be the space of gauge freedoms. If $\Lambda$ orthogonalizes $\Theta$ and $G$, then $\Lambda$ is a $\Theta$-regularizer.
\end{lemma}
\noindent Moreover, Posfai et al.~establish two conditions that are equivalent to $\Lambda$ orthogonalizing a pair of complementary subspaces $V_1$ and $V_2$; we restate this result as \Cref{lem:gauge-claim-25} in \Cref{sec:useful}.
In the following lemma, we establish another equivalent condition that we will use in the proof of \Cref{thm:alway-can-regularize}, see \Cref{sec:useful} for the proof.

\begin{lemma}\label{lem:form-of-Lambda} Let $V_1$ and $V_2$ be complementary subspaces. A matrix $\Lambda$ orthogonalizes $V_1$ and $V_2$ if and only if $\Lambda = A^TA + B^TB$ for matrices $A$ and $B$ such that $\nullspace(A) = V_2$ and $\nullspace(B)= V_1$. 
\end{lemma}

 \subsection{Equivalence of MAP estimates and optimizers of regularized regression}\label{subsec:equiv}

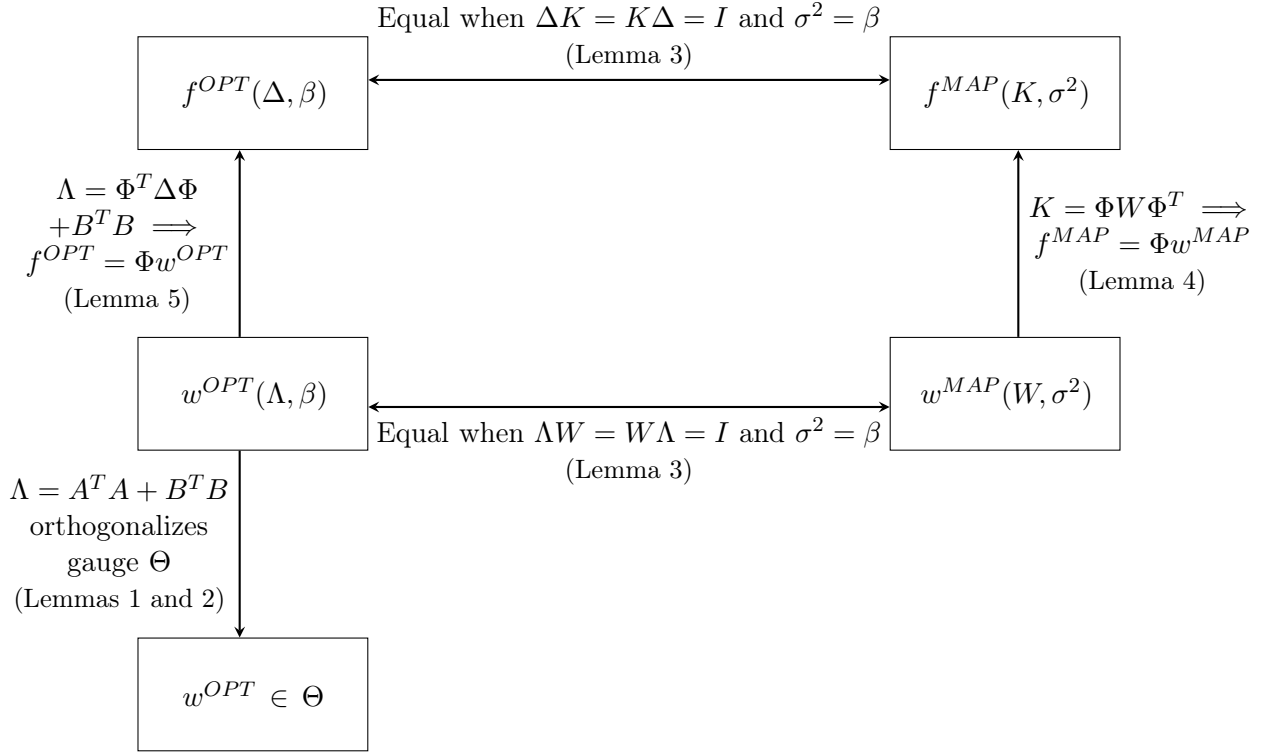
\begin{figure}[h]
\begin{tikzpicture}[>=stealth]
    \node[draw, text width=2.8cm, minimum height=1.5cm, align=center] (box1) at (0,0) {$f^{OPT}(\Delta, \beta)$};
    \node[draw, text width=2.8cm, minimum height=1.5cm, align=center] (box2) at (10,0) {$f^{MAP}(K, \sigma_n^2)$};
    \node[draw, text width=2.8cm, minimum height=1.5cm, align=center] (box3) at (0,-4) {$w^{OPT}(\Lambda, \beta)$};
    \node[draw, text width=2.8cm, minimum height=1.5cm, align=center] (box4) at (10,-4) {$w^{MAP}(W, \sigma_n^2)$};
    \node[draw, text width=2.8cm, minimum height=1.5cm, align=center] (box5) at (0,-8) {$w^{OPT} \in \Theta$};

    \draw[<->, thick] ([yshift=5pt] box1.east) -- node[above, align=center] {Equal when $\Delta K = K\Delta = I$ and $\sigma_n^2 = \beta$ \\ \small{(\Cref{lem:gp-map-equiv})}} ([yshift=5pt] box2.west);
    \draw[<-, thick] ([xshift=-5pt] box1.south) -- node[left, align=center] {$\Lambda = \Phi^T\Delta \Phi$\\  $+B^TB \implies$\\$f^{OPT}= \Phi w^{OPT}$\\ \small{(\Cref{lem:form-of-regularizer-prior-gauge})}} ([xshift=-5pt] box3.north);
    \draw[<-, thick] ([xshift=5pt] box2.south) -- node[right,align=center] {$K=\Phi W\Phi^T \implies$\\ $ f^{MAP}= \Phi w^{MAP}$\\\small{(\Cref{lem:gp-equiv-coeff-func})}} ([xshift=5pt] box4.north);
    \draw[<->, thick] ([yshift=-5pt] box3.east) -- node[below, align=center] {Equal when $\Lambda W = W \Lambda = I$ and $\sigma_n^2 = \beta$\\\small{(\Cref{lem:gp-map-equiv})}} ([yshift=-5pt] box4.west);
    \draw[->, thick] ([xshift=-5pt] box3.south) -- node[left, align=center] {$\Lambda= A^TA+B^TB$\\  orthogonalizes\\  gauge $\Theta$\\ \small{(\Cref{lem:form-of-Lambda,lem:orthogonalizes-implies-opt-in-gauge})}} ([xshift=-5pt] box5.north);
\end{tikzpicture}
\caption{An illustration of the equivalences established. Let $A$ and $B$ be matrices such that $nullspace(A) = G$ and $nullspace(B)= \Theta$. }
    \label{fig:boxes}
\end{figure}

In this section we will establish the equivalence of the following prediction methods:
  \begin{enumerate}
     \item \textbf{Penalized regression in weight space} (\Cref{defn:wopt}). 
     
     $$w^{OPT}(\Lambda, \beta) = \argmin_{w \in \R^{(\alpha+1)^\ell}} \| y-\Phi_Xw\|_2^2 + \beta w^T \Lambda w$$
     \item \textbf{Gaussian process regression in function space} (\Cref{defn:fmap}). Suppose $y = f_X +\ve$ where $f \sim N(0,K)$ and $\ve \sim N(0, \sigma_n^2 I)$. Let $f^{MAP}(K, \sigma_n^2)$ be the MAP estimate for $f$ under this Gaussian process.
 \end{enumerate}
 To do so, we introduce two additional prediction methods as intermediaries and establish the equivalences depicted in \Cref{fig:boxes}. 
 \begin{enumerate}
  \setcounter{enumi}{2}  
    \item \textbf{Penalized regression in function space.}
    
 $$f^{OPT}(\Delta, \beta) = \argmin_{f \in \R^{\alpha^\ell}} \| y-f_X\|_2^2 + \beta f^T \Delta f$$
    \item \textbf{Bayesian regression in weight space.} Suppose $y = \Phi_X w + \ve$ where $w \sim N(0,W)$ and $\ve \sim N(0, \sigma_n^2 I)$. Let $w^{MAP}(W, \sigma_n^2)$ be the MAP estimate for $w$ under this Gaussian prior. 
 \end{enumerate}

 \noindent With respect to the last of these methods, it is well-known that the posterior distribution of $w$ is 
 
\begin{equation}\label{eq:ws-prior}
    N(w^{MAP}, A^{-1}) \quad \text{ where }\quad w^{MAP} = \sigma^{-2} A^{-1} \Phi_X^T y \quad \text{ and } \quad A = \sigma^{-2} \Phi_X^T \Phi_X + W^{-1}.
\end{equation}

We begin with two well-known lemmas. \Cref{lem:gp-map-equiv} relates the optimizer of penalized regression and the MAP estimate of a Gaussian process (see Section 6.2 of \cite{Rasmussen06}), and  \Cref{lem:gp-equiv-coeff-func} establishes the equivalence of the MAP estimates for Bayesian regression in weight space and Gaussian process regression in function space (see Section 2.1.2 of \cite{Rasmussen06}).

\begin{lemma}\label{lem:gp-map-equiv} Suppose $y = \Phi_Xb + \ve$ where $b \sim N(0,W)$ and $\ve \sim N(0, \sigma_n^2 I)$. Then the MAP estimate for $b$ is equal to 

$$\argmin_{b} \| y-\Phi_Xb\|_2^2 + \sigma_n^2 b^T W^{-1} b.$$
\end{lemma}

\begin{lemma}\label{lem:gp-equiv-coeff-func}
Let $X,y$ be observations, $\sigma_n^2$ be the noise variance, and $K$ and $W$ be kernels such that $K = \Phi W\Phi^T$. Let $f$ be drawn from the posterior distribution of the function space Gaussian process defined by $K, \sigma_n^2, X,y$ (\Cref{eq:fs-prior}), and $w$ be drawn from the posterior distribution of Bayesian regression in weight space defined by $W, \sigma_n^2, X,y$ (\Cref{eq:ws-prior}). Then the posterior distribution of $\Phi w $ is identical to the posterior distribution of $f$.
Thus, $f^{MAP}(K, \sigma_n^2)= \Phi w^{MAP}(W, \sigma_n^2)$.
 \end{lemma}

The following lemma establishes equivalence of the optimizers for regularized regression in weight space and function space.

\begin{lemma}\label{lem:form-of-regularizer-prior-gauge} Let $B$ be a matrix with nullspace  $\Theta$, and let $\Delta$ be positive-definite.  If $\Lambda = \Phi^T \Delta \Phi + B^TB$ then $f^{OPT}(\Delta, \beta) = \Phi w^{OPT}(\Lambda, \beta)$ for all $\beta>0$.    
\end{lemma}

\begin{proof} 
We will use two facts. 
First, suppose $w \in \Theta$. Then 

$$ w^T \Lambda w = w^T(\Phi^T\Delta \Phi + B^TB)w = (\Phi w)^T \Delta (\Phi w).$$
Second,  $w^{OPT} \in \Theta$. This follows directly from \Cref{claim:phi-delta-phi}, which states that $\Phi^T \Delta \Phi = A^TA$ for some $A$ with null space equal to the space of gauge freedoms $G$, and \Cref{lem:form-of-Lambda,lem:orthogonalizes-implies-opt-in-gauge}.

Let $w_0$ be the representation of $f^{OPT}$ in gauge $\Theta$,
$\Phi w_0=f^{OPT}$. By the first fact, $\brac{f^{OPT}}^T \Delta f^{OPT} = (\Phi w_0)^T \Delta (\Phi w_0) = w_0^T \Lambda w_0$. It follows that 
 
 $$\psi = \| y-\Phi_Xw_0\|_2^2 + \beta w_0^T \Lambda w_0=\|y - f^{OPT}_X\|_2^2 + \beta \brac{f^{OPT}}^T \Delta f^{OPT}. $$
The optimality of $w^{OPT}$ and $f^{OPT}$ imply that 
$$ \| y-\Phi_Xw^{OPT}\|_2^2 + \beta \brac{w^{OPT}}^T \Lambda w^{OPT}  \leq \psi \leq \|y - \brac{\Phi w^{OPT}}_X\|_2^2 + \beta \brac{\Phi w^{OPT}}^T \Delta \brac{\Phi w^{OPT}}. $$
Note that $\| y-\Phi_Xw^{OPT}\|_2^2= \|y - \brac{\Phi w^{OPT}}_X\|_2^2 $. The fact that $w^{OPT} \in \Theta$ implies $\brac{\Phi w^{OPT}}^T \Delta \brac{\Phi w^{OPT}} = \brac{w^{OPT}}^T \Lambda w^{OPT}$. Thus the upper and lower bounds are equal meaning that inequalities are satisfied at equality. Thus we have 
$$\psi =\| y-\Phi_Xw_0\|_2^2 + \beta w_0^T \Lambda w_0  =  \| y-\Phi_Xw^{OPT}\|_2^2 + \beta \brac{w^{OPT}}^T \Lambda w^{OPT},$$ and the uniqueness the optimizer implies that $w^{OPT} = w_0$ meaning $f^{OPT} = \Phi w^{OPT}$, as desired.
\end{proof}

\subsection{Proof of \Cref{thm:alway-can-regularize} and Bayesian regression form of  \Cref{thm:alway-can-regularize}  }\label{sec:br-form}

With the results in the previous sections, we now prove \Cref{thm:alway-can-regularize}, which describes how to build a $\Theta$-regularizer that induces a function space prior $K$. 
\begin{proof} (of \Cref{thm:alway-can-regularize})  Take $\Lambda =\Phi^T K^{-1} \Phi  + B^TB$ where $B$ is a matrix with nullspace $\Theta$. \Cref{claim:phi-delta-phi} and \Cref{lem:form-of-Lambda,lem:orthogonalizes-implies-opt-in-gauge} directly imply that $\Lambda$ is a $\Theta$-regularizer. The fact that $\Lambda$ induces the prior $K$  follows directly from \Cref{lem:gp-map-equiv,lem:form-of-regularizer-prior-gauge}.
\end{proof}

The inverse of a $\Theta$-regularizer that induces function space prior $K$ is a Bayesian regression prior $W$ that yields an MAP estimate in $\Theta$ and induces the prior $K$, meaning $w \sim N(0, W)$ implies $\Phi w \sim N(0,K)$.
\begin{remark}
    \label{remark:b-regression-form}
     For any linear gauge $\Theta$ and prior on function space $K$, there exists a Gaussian prior over weight space $w\sim N(0, W)$ such that $\Phi w\sim N(0, K)$ and $w^{MAP}(W, \sigma_n^2) \in \Theta.$ The matrix $W=\brac{\Phi^T K^{-1} \Phi  + B^TB}^{-1}$ where $B$ is a matrix with nullspace $\Theta$ is one such prior. 
\end{remark}
\noindent Note \Cref{lem:gp-map-equiv} implies $w^{OPT}(W^{-1}, \sigma_n^2)= w^{MAP}(W, \sigma_n^2)$, and \Cref{thm:alway-can-regularize} establishes that $w^{OPT} \in \Theta$. To see that $\Phi w\sim N(0, K)$, we compute the probability density for the event that $\Phi w = f$ for a fixed $f \in \R^{\alpha^\ell}$. Let $\theta$ be the gauged-fixed representation of $f$. Whenever $\Phi w =f$,  $w = \theta + g$ where $g$ is in the space of gauge freedoms. 
Thus for $w \sim N(0,W)$, the probability density of the event that  $\Phi w = f $ is equal to  \begin{align*} 
c \int_G \exp{ -\frac{1}{2} (\theta + g)^T \brac{\Phi^T K^{-1} \Phi  + B^TB}(\theta + g) } dg &= c' \exp{ -\frac{1}{2}f^TK^{-1}f}, \end{align*}
for constants $c$ and  $c'$ that do not depend on $w$ (with $c' = c\int_G \exp{ -\frac{1}{2} g^T B^TBg } dg$), and where we have used the fact that $g$ is in the nullspace of $\Phi$ and $\theta$ is in the nullspace of $B$. 

\section{Building gauge-specific regularizers that induce variance component and product kernel priors}\label{sec:build-it}
In this section we demonstrate how to build $\Theta$-regularizers that induce function space priors $K$ for a useful class of gauges $\Theta$ and two useful classes of function space priors $K$. 
Recall that for a matrix $B$ with nullspace $\Theta$, 

$$\Lambda =\Phi^T K^{-1} \Phi  + B^TB$$ is a $\Theta$-regularizer that induces prior $K$. This additive form gives us a recipe for building $\Theta$-regularizers that induce a specific function space prior $K$, where the regularizer takes the form of  a sum of two matrices, one that determines the induced prior and the other that determines the gauge. In \Cref{sec:gauges}, we derive a simple formula for a matrix of the form $B^TB$ where the nullspace of $B$ is $\Theta^{\lambda,\pi}$. Then in \Cref{sec:priors-in-reg} we compute  $\Phi^T K^{-1} \Phi$ for two classes of kernels: variance component kernels and product kernels.  Together these results serve as a recipe for constructing regularizers that induce a particular variance component or product prior in a particular $\lambda$-$\pi$ gauge.

\subsection{Building $\Theta$-regularizers for $\lambda$-$\pi$ gauges}\label{sec:gauges}

In \Cref{subsec:marg-prop} we establish a marginalization property for $\lambda$-$\pi$ gauges that helps to interpret the gauge parameters. Then in \Cref{subsec:build-btb} we employ this marginalization property to build simple matrices of the form $B^TB$ where the nullspace of $B$ is a $\lambda$-$\pi$ gauge $\Theta^{\lambda,\pi}$. 

\subsubsection{A marginalization property for $\lambda$-$\pi$ gauges}\label{subsec:marg-prop}

\begin{definition} We say that a vector $w \in \R^{(\alpha+1)^\ell}$ satisfies the $\lambda$-$\pi$ marginalization property if for all $(U,u,p)$ where $U$ is a subset of positions that does not contain $p$ and $u$ is a subsequence on $U$, 
\begin{equation}\label{eq:marginalization}
    \sum_{c \in \mathcal{A}} \pi^p_c w_{(U \cup \{p\}, u^{+c})} = \frac{w_{(U,u)}}{\lambda}=\bfrac{1-\eta}{\eta} w_{(U,u)},
\end{equation}
where $\eta = \lambda/(1+\lambda)$, and  $u^{+c}$ denotes the sequence on $U \cup \{p\}$ that agrees with $u$ on $U$ and has character $c$ at position $p$.  
\end{definition}

\Cref{lem:marg-property} establishes that $w$ satisfies the $\lambda$-$\pi$ marginalization property if and only if $w$ is in the $\lambda$-$\pi$ gauge. This marginalization property was established for the hierarchical gauge ($\lambda = \infty, \eta =1$) in Claim 21 of \cite{posfai2024gauge}. In the zero-sum gauge (hierarchical gauge with uniform $\pi$), the weights for any set of $\alpha$ subsequences on the same set of positions that differ only at one particular position average to zero.  Direct consequences of this are the interpretation of the weights as mean effects and the fact that the truncated model (the function obtained by summing weights on subsequences of length up to $k$) is the least-squares approximation of the function with up to $k^{th}$ order terms, as described in \Cref{subsec:lam-pi-fam}. For a non-uniform product distribution $\pi$, the results are analogous with a weighted average and weighted least squares with respect to $\pi$.

Our marginalization property is novel for finite $\lambda$, and its interpretation provides insight into how $\lambda$ affects magnitude of the weights as a function of the size of the corresponding subsequence. The hierarchical ($\lambda = \infty$) case can be thought of as pushing as much information into the lower order weights as possible; the best $k^{th}$ order approximation can be obtained by truncating up to weights corresponding to subsequences of length $k$. Consider the other extreme when $\lambda$ is very small.  The weights for any set of $\alpha$ subsequences on the same set of $k$ positions that differ only at one particular position average to $1/\lambda$ times the weight for the length $k-1$ subsequence in common, yielding relatively larger weights for longer subsequences. As $\lambda \to 0$, all except the highest order weights (corresponding to length $\ell$ subsequences) vanish, yielding the trivial gauge.  

\begin{lemma}\label{lem:marg-property} A vector of weights $w$ satisfies the $\lambda$-$\pi$ marginalization property if and only if $w \in \Theta^{\lambda, \pi}$.
\end{lemma}
\begin{proof}
First we show that if $w \in \Theta^{\lambda,\pi}$, then $w$ satisfies the $\lambda$-$\pi$ marginalization property. It suffices to fix a basis for $\Theta^{\lambda, \pi}$  and show that each basis vector satisfies the $\lambda$-$\pi$ marginalization property. 
    We can choose a basis for $\Theta^{\lambda, \pi}$ where each basis vector has the form $w = \bigotimes_{p=1}^\ell \theta^p$ where $\theta^p \in \Theta^{\lambda, \pi^p}$. Recall the construction of $\Theta^{\lambda, \pi^p}$ given in \Cref{eq:building-blocks}.  Note $\theta^p$ is an $\alpha +1$ dimensional vector. We index the last $\alpha$ positions with the corresponding character $c$ and index the first position with $0$. Since only the $V_\lambda$ component of $w$ contributes to a nonzero value to $\theta_0$, 
    
    $$\theta^p - \frac{\theta^p_0}{\lambda }  \begin{pmatrix}
\lambda \\
1 \\
\vdots \\
1
\end{pmatrix} \in V_\perp^{\pi^p},\quad \quad \text{
so } \sum_{c \in \mathcal{A}} \pi^p_c \brac{ \theta^p_c - \frac{\theta^p_0}{\lambda}} =0 \quad \text{ and thus } \quad \sum_{c \in \mathcal{A}} \pi^p_c \theta_c^p= \frac{\theta^p_0}{\lambda}.$$

 Let $U$ be a subset of positions that does not contain $p$ and $u$ be a subsequence on $U$. We now show that the basis element $w$ satisfies \Cref{eq:marginalization}. Note 

$$w_{(U \cup \{p\}, u^{+c})} = \zeta \theta^p_c \quad \text{ and } \quad w_{(U,u)} = \zeta \theta^p_0 \quad \text{ where } \quad \zeta = \prod_{q \in U}\theta^q_{u_q} \prod_{q \not \in U \cup \{p\}} \theta^q_0.$$
It follows that  

$$\sum_{c \in \mathcal{A}} \pi^p_c w_{(U \cup \{p\}, u^{+c})} = \zeta\sum_{c \in \mathcal{A}} \pi^p_c \theta^p_c =\frac{\zeta \theta^p_0}{\lambda} = \frac{w_{(U,u)}}{\lambda},$$
as desired.

Next we show that any $w$ that satisfies the $\lambda$-$\pi$ marginalization property is in $\Theta^{\lambda, \pi}$ by showing that $P^{\lambda, \pi} w=w$. Recall the expression for the projection matrix given in \Cref{eq:proj}.  Let $(S,s)$ be an arbitrary subsequence. We will show that $(P^{\lambda, \pi} w)_{(S,s)}=w_{(S,s)}$, or equivalently 

$$\sum_{(T,t)} b(T,t) w_{(T,t)} = w_{(S,s)},$$
where for ease of notation we let 
$$b(T,t) = P^{\lambda, \pi}_{(S,s), (T,t)} = 
\prod_{p \in S\cap T}\brac{\delta_{s_p = t_p} -\pi^p_{t_p} \eta} \prod_{p \in S \setminus T} (1- \eta) \prod_{p \in T \setminus S}\pi^p_{t_p} \eta  \prod_{p \not \in S\cup T} \eta. $$

Let $F$ be a subset of positions, and let $\mathcal{T}_F$ be the set of subsequences that agree with $s$ on positions in $S \cap F$ and do not include positions in $S^c \cap F$,

$$\mathcal{T}_F = \{ (T,t):  t_p=s_p \text{ for all } p \in S \cap F \cap T, \text{ and }  p \not \in T \text{ for all } p \in S^c \cap F \}.$$

We iterative apply \Cref{claim:p-in-S,claim:p-not-in-S} (proved in \Cref{sec:marg-help-results}) for each position $p \in S$ and each position $p \not \in S$ respectively to obtain the desired result
\begin{align*}
    \sum_{(T,t) \in \mathcal{T}_\emptyset} b(T,t) w_{(T,t)}&= \brac{\prod_{p \in S} \frac{1}{1-\pi^p_{s_p}\eta} \prod_{p \not \in S} \frac{1}{\eta}} \sum_{(T,t) \in \mathcal{T}_{\{1,2, \dots \ell\}}} b(T,t) w_{(T,t)}\\
    &=\brac{ \prod_{p \in S} \frac{1}{1-\pi^p_{s_p}\eta} \prod_{p \not \in S} \frac{1}{\eta} }b(S,s) w_{(S,s)}=w_{(S,s)}.
\end{align*}

\end{proof}

\subsubsection{A simple formula for $B^TB$ for $\Theta^{\lambda, \pi}$ regularizers }\label{subsec:build-btb}

We derive a simple formula for $B^TB$ for $\Theta^{\lambda, \pi}$ regularizers that is as sparse as the Laplacian of the Hamming graph for words of length $\ell$ with alphabet size $\alpha +1$.

\begin{theorem}
Let 

$$Z_{(S,s),(T,t)}=\begin{cases}
    \bfrac{1-\eta}{\eta}^2 + \sum_{p \in S} \brac{\pi^p_{s_p}}^2 & S=T,\, s=t\\
    \pi^p_{s_p} \pi^p_{t_p} & S=T, \,d(s,t)= 1 \text{ with } s_p \not = t_p\\
    0& \text{otherwise}
\end{cases}.$$
Then $Z= B^TB$ for a matrix $B$ with null space equal to the gauge space $\Theta^{\lambda, \pi}$.
\end{theorem}

\begin{proof} Let $B$ be an $\ell(\alpha+1)^{\ell-1} \times (\alpha +1)^\ell$ matrix with columns indexed by subsequences $(T,t)$ and rows indexed by triples $(U,u,p)$ where $U$ is a subset of positions that does not contain $p$ and $u$ is a subsequence on $U$. Recall that $u^{+c}$ denotes the sequence on $U \cup \{p\}$ that agrees with $u$ on $U$ and has character $c$ at position $p$. Define

$$B_{(U,u,p),(T,t)}= \begin{cases}
\pi^p_c & T= U \cup \{p\}, t= u^{+c}\\
\frac{-(1-\eta)}{\eta} & T=U, t=u \\
0& otherwise,
\end{cases}.$$
where $\eta = \lambda/(1+\lambda)$. 
Note that $w \in \nullspace(B)$ is equivalent to $w$ satisfying the $\lambda$-$\pi$ property. \Cref{lem:marg-property} implies that $\Theta^{\lambda, \pi} = \nullspace(B)$.
It is straightforward to verify that $Z=B^TB$.
\end{proof}

\subsection{Variance component and product kernels}\label{sec:kern-defn}

Here we introduce two classes of kernels that have shown to perform well in practice: variance component kernels \cite{zhou2022higher} and product kernels \cite{zhou2024tbd}.

 Variance component (VC) kernels are the class of isotropic kernels, i.e.~kernels whose values depend only on the Hamming distance between the pair of sequences. These kernels are parameterized in terms of how much variance is due to different orders of interaction. We can decompose function space into orthogonal subspaces $V_0, V_1,  \dots V_{\ell}$, where $V_0$ is the constant subspace and each $V_k$ includes function vectors that express interactions between exactly $k$ sites; for further details see \cite{zhou2022higher, neidhart2013exact, happel1996canonical}. Consequently we can decompose any vector uniquely as the sum of orthogonal vectors $f = \sum_{k=0}^\ell f_k$ where $f_k \in V_k$, and decompose the variance  $f^Tf = \sum_{k=0}^\ell f_k^T f_k$. For $f$ drawn from a mean zero distribution, we define the \textit{variance of order $k$} of this distribution as $\E{f_k^T f_k}$. Krawtchouk polynomials are used to build kernels that are parameterized by these variances. 

\begin{definition}\label{defn:kraw}
    The Krawtchouk polynomial is
    
    $$\mathcal{K}_k(d) = \mathcal{K}_k(d;\ell, \alpha)=\sum_{i =0}^{k} (-1)^i (\alpha-1)^{k-i} {d \choose i} {\ell-d \choose k-i}.$$
\end{definition}

\begin{definition} Let $\lambda_0, \dots, \lambda_\ell>0$. The variance component (VC) kernel is given by 

$$K_{x,y} = \sum_{k=0}^\ell \lambda_k \mathcal{K}_k(d(x,y))$$
where $d(x,y)$ denotes the Hamming distance between sequences $x$ and $y$.
\end{definition}
For $f$ drawn from a mean zero VC kernel, $\E{f_k^T f_k} =\lambda_k {\ell \choose k} (\alpha-1)^k$. The dimension of $V_k$ is ${\ell \choose k} (\alpha-1)^k$, so we call $\lambda_k$ the \textit{dimension-normalized variance of order $k$}. Note that $\lambda_k$ can be interpreted as the mean squared interaction coefficients of order $k$, see \cite{zhou2022higher}. 

We will also consider product kernels, which are not necessarily isotropic priors that can express the fact that different position-character combinations can play different roles. We recently introduced two subclasses of product kernels, connectedness and Jenga kernels, that achieve state-of-the-art predictive performance on several sequence-function datasets and yield hyperparameters that can be interpreted to provide insight into the mechanisms by which the sequence determines the function \cite{zhou2024tbd}. We formally define these kernels in \Cref{sec:defn-jenga}. 

\begin{definition} A product kernel on sequences of length $\ell$ has the form 

$$K = \bigotimes_{p=1}^\ell K^p,$$
    where for $p \in [\ell]$, $K^p$ is a symmetric $\alpha \times \alpha$ positive-definite matrix. 
\end{definition}

\subsection{Building regularizers that induce VC and product kernels. }\label{sec:priors-in-reg}
First we compute $\Phi^T K^{-1} \Phi$ when $K$ is a VC kernel. While this matrix is dense, it contains only order $\ell^2$ distinct entries which can be precomputed. 
\begin{theorem}\label{thm:phi-kinv-phi-vc}
    Let $K$ be a VC kernel,  $K_{x,y} = \sum_{k=0}^\ell \lambda_k \mathcal{K}_k(d(x,y))$, then $$(\Phi^T K^{-1} \Phi)_{(S,s),(T,t)} =  \alpha^{\ell-|S\cup T|} \sum_{d=j}^{\ell- (|S\cap T| -j)} {\ell - |S \cap T| \choose d-j} (\alpha -1)^{d-j}\brac{\sum_{k=0}^\ell \lambda_k^{-1} \mathcal{K}_k(d)},$$
   where  $j$ is the Hamming distance between $s$ and $t$ among their common positions $S \cap T$.
\end{theorem}

\begin{proof}
Note

$$\brac{\Phi^T K^{-1} \Phi}_{(S,s),(T,t)}=  \sum_x \Phi_{x,(S,s)} \brac{\sum_y K^{-1}_{x,y} \Phi_{y,(T,t)}} = \sum_{\substack{x,y\\x[S]=s, y[T]=t}} K^{-1}_{x,y}.$$
To compute this sum, we count how many pairs of sequences $x,y$ are at distance $d$ and satisfy $x[S]=s, y[T]=t$. In building such pairs, the subsequence of $x$ on $S$ is fixed and the subsequence of $y$ on $T$ is fixed. Let $j$ be the Hamming distance between $s$ and $t$ among their common positions $S \cap T$. All valid pairs of sequences will have Hamming distance at least $j$.
There are $\alpha^{\ell- |S\cup T|}$ choices for the subsequence of $x$ on $(S \cup T )^c$. Having fixed one such subsequence on these positions in $x$, each position in $(S \cap T)^c$ is fixed in either $x$ or $y$ (in $S \setminus T$, $x$ is fixed; in $T \setminus S$, $y$ is fixed; in $(S\cup T)^c$, $x$ is fixed). To build a pair of sequences at distance $d$, we need to choose $d-j$ positions from $(S \cap T)^c$ and pick characters so that $x$ and $y$ disagree at those positions. There are ${\ell-|S\cap T| \choose d-j}(\alpha-1)^{d-j}$ ways to do so.  We apply \Cref{lem:vc-inverse} to compute $K^{-1}$ and obtain

$$\brac{\Phi^T K^{-1} \Phi}_{(S,s),(T,t)}= \sum_{d=j}^{\ell- (|S\cap T| -j)}  \alpha^{\ell-|S\cup T|} {\ell - |S \cap T| \choose d-j} (\alpha -1)^{d-j}\brac{\sum_{k=0}^\ell \lambda_k^{-1} \mathcal{K}_k(d)}.$$
\end{proof}

Next we compute $\Phi^T K^{-1} \Phi$ for product kernels; this matrix is again dense, but it can easily be computed as the tensor product of $\ell$ matrices with simple forms. From \Cref{thm:phi-kinv-phi-product}, it is straightforward to compute $\Phi^TK^{-1}\Phi$ for Jenga, connectedness, and Geometric kernels. For completeness we include these computations in \Cref{sec:jenga-cors}.

\begin{theorem}\label{thm:phi-kinv-phi-product}
    Let $K$ be a product kernel with $\brac{K^p}^{-1}_{c,c'}= b^p_{c,c'}$. Then 
    
    $$\brac{\Phi^T K^{-1} \Phi}_{(S,s),(T,t)}= \prod_{p \in S\cap T} b^p_{x_p,y_p} \prod_{p \in S \setminus T} \brac{\sum_{c \in \mathcal{A}} b^p_{s_p,c}} \prod_{p \in T \setminus S} \brac{\sum_{c \in \mathcal{A}} b^p_{t_p,c}} \prod_{p \not \in S \cup T} \brac{\sum_{c,c'} b^p_{c,c'}}$$
\end{theorem}

\begin{proof} Since $K^{-1} = \bigotimes_p (K^{p})^{-1}$, $K^{-1}_{x,y} = \prod_p b^p_{x_p,y_p}$. Recall $\Phi_{X,(S,s)} = \delta_{x[S]=s}$. It follows that

$$\brac{\Phi^T K^{-1} \Phi}_{(S,s),(T,t)}=\sum_{\substack{x,y\\x[S]=s, y[T]=t}} K^{-1}_{x,y}= \sum_{\substack{x,y\\x[S]=s, y[T]=t}} \prod_p b^p_{x_p,y_p}.$$
 We construct pairs of sequences $x$ and $y$ for which $x[S]=s, y[T]=t$ and compute the factor that each position contributes to the summand.
\begin{itemize}
    \item If $p \in S\cap T$, there is only one option: $x_p=s_p$ and $y_p=t_p$, and this position contributes a factor of $b^p_{x_p,y_p}$ to the summand.
    \item Consider $p \in S\setminus T$. We must have $x_p=s_p$ and $y_p$ can take any value. If $y_p=c$, this position contributes a factor of $b^p_{x_p,c}$ to the summand. The case $p \in T \setminus S$ is analogous. 
    \item Consider $p \not \in S \cup T$. Then $x_p$ and $y_p$ can take any value. If $x_p=c$ and $y_p = c'$, this position contributes a factor of $b^p_{c,c'}$ to the summand.
  
\end{itemize}
Note that each term in the expansion of the following expression corresponds to the $K^{-1}_{x,y}$ value for a pair of sequences with the property that $x[S]=s, y[T]=t$. The result follows.

$$ \prod_{p \in S \cap T}  b^p_{x_p,y_p}
 \prod_{p \in S \setminus T} \brac{\sum_{c \in \mathcal{A}} b^p_{x_p,c}}
  \prod_{p \in T \setminus S} \brac{\sum_{c \in \mathcal{A}} b^p_{y_p,c}}
 \prod_{p \not \in S \cup T} \brac{\sum_{c,c'} b^p_{c,c'} }.$$

\end{proof}

\section{Function space priors induced by diagonal regularizers}\label{sec:diag}

A natural choice of a weight space regularizer is a diagonal matrix $\Lambda$. The optimizer of weight space regression with a diagonal regularizer equals the MAP estimate under a Gaussian prior for weight space in which the weights are assumed to be drawn independently. There is a large literature on random fitness landscapes constructed as the weighted sum of subsequence indicator features where the weights are drawn independently, see \cite{kauffman1989nk,perelson1995protein,buzas2013analysis,hwang2018universality}. One such model is the GNK (generalized NK model \cite{buzas2013analysis,nowak2015analysis,hwang2018universality, brookes2022sparsity}). In the GNK model, each position $p$ defines a subset of positions $N_p$ with $p \in N_p$ called a neighborhood. Weights for subsequences on the neighborhoods, $(N_p,s)$, are drawn independently from normal distributions whose variance is the inverse of the size of $N_p$ and zero weight is assigned to all other subsequences. Therefore, the GNK model is induced by a diagonal regularizer of the form $\Lambda_{(S,s),(S,s)} =  |S|$ when $S$ is a neighborhood $S=N_p$ and $\infty$ when $S$ is not a neighborhood.

Here we compute the function space priors induced by diagonal regularizers in which the regularization strength is finite for all weights. In \Cref{subsec:diag-for-lam-pi}, we show that the class of $\Theta^{\lambda, \pi}$ diagonal regularizers (for $\lambda< \infty$) introduced in \cite{posfai2024gauge} induce product kernel priors on function space. Then in \Cref{sec:gen-diag}, we consider the class of diagonal regularizers where the regularization strength depends only on the order of interaction. We show that such regularizers induce VC priors on function space, but not all VC kernels can be induced by these diagonal regularizers. 

\subsection{Function space priors induced by diagonal $\Theta^{\lambda,\pi}$-regularizers}\label{subsec:diag-for-lam-pi}

Diagonal weight space regularizers of the form

$$\Lambda_{(S,s),(S,s)} = \lambda^{|S|} \prod_{p \in S} \pi^p_{s_p}$$ are $\Theta^{\lambda, \pi}$-regularizers \cite{posfai2024gauge}. This form emphasizes the relationship between $\lambda$ and the distribution of weights across different subsequence lengths. With the uniform distribution $\pi$, $\Lambda_{(S,s),(S,s)} =(\lambda /\alpha)^{|S|}$. When $\lambda = \alpha$ all weights incur equal penalization. Larger $\lambda$ disproportionately penalizes the weights for higher-order subsequences, whereas smaller $\lambda$ disproportionately penalizes the weights for lower-order subsequences. 
Here we compute the function space prior induced by such regularizers.

\begin{theorem}
Let $\lambda >0$ be finite and $\pi$ a product distribution with full support.  The diagonal regularizer $\Lambda_{(S,s),(S,s)} = \lambda^{|S|} \prod_{p \in S} \pi^p_{s_p} $ induces the prior 

$$K_{x,y}= \prod_{p : x_p = y_p} \brac{ 1 + \frac{1}{\pi^p_{x_p} \lambda}}.$$
\end{theorem}

\begin{proof}
    By \Cref{lem:gp-map-equiv,lem:gp-equiv-coeff-func}, it suffices to compute $\Phi \Lambda^{-1} \Phi^T$,
    
    $$
    (\Phi \Lambda^{-1} \Phi^T)_{x,y}   = \sum_{\substack{S,s:\\ x[S]=s,\, y[S]=s}}\Lambda^{-1}_{S,s,T,t}
    = \sum_{\substack{S,s:\\ x[S]=s,\, y[S]=s}}\frac{1}{\lambda^{|S|} \prod_{p \in S} \pi^p_{s_p}} = \prod_{p : x_p = y_p} \brac{ 1 + \frac{1}{\pi^p_{x_p} \lambda}}.
$$
\end{proof}

Note that for non-uniform $\pi$, this is a heteroskedastic prior; the variances for sequences more likely under $\pi$ are comparatively smaller. 
For uniform $\pi$, the induced kernels are scaled geometric decay kernels with $\beta = \brac{1 + \frac{\alpha}{\lambda}}^{-1}$ and scale factor 
$\brac{1 + \frac{\alpha}{\lambda}}^\ell$,

$$K_{x,y} =\brac{1 + \frac{\alpha}{\lambda}}^{\ell - d(x,y)}$$
When $\lambda$ is smaller, we observe a sharper decay in correlation. This aligns with the observation that when $\lambda$ is smaller the diagonal regularizer penalizes the weights for higher-order subsequences less strongly; when weights of higher-order subsequences are free to vary widely, the resultant function will have less correlation across similar sequences. 

\subsection{Function space priors induced by order-dependent diagonal regularizers}\label{sec:gen-diag}

Next we consider diagonal regularizers with values depending only on the length of the subsequences.

\begin{theorem}\label{thm:prior-from-diag-reg} Let $\Lambda$ be a diagonal regularization matrix whose values depends only on the length of the subsequence,

$$\Lambda_{(S,s),(S,s)} = a_{|S|}, $$ where $a_j >0$ for $j \in [\ell]$. Then $\Lambda$ induces the VC prior 

$$K_{x,y} =    \sum_{k=0}^\ell \brac{\sum_{j=k}^\ell \frac{1}{\alpha^j a_j}  {\ell-k \choose j-k}}\mathcal{K}_k(d(x,y)).$$
\end{theorem}

While any sequence of positive $\lambda_k$'s defines a valid VC kernel, we show in \Cref{sec:cannot-achieve} that not all VC kernels are induced by some order-dependent diagonal regularizer. 
The following corollary explains how to compute a order-dependent regularizer that induces a VC kernel, if one exists.

\begin{corollary}
    Let $K$ be the kernel defined by dimension-normalized variance components $\lambda\geq 0$.  Let $T$ be an $(\ell+1) \times (\ell+1)$ zero-indexed upper triangular matrix with $T_{ij} = {\ell-i \choose j}$ for $i\leq j$. Let  $W$ be an $(\ell+1) \times (\ell+1)$ zero-indexed matrix with $W_{ij}= w_j(i)$. If $a= T^{-1} W \lambda >0$ entry-wise, then the order-dependent diagonal regularizer  $\Lambda_{(S,s),(S,s)} = 1/a_{|S|}$ induces the prior $K$. 
\end{corollary}

We now prove \Cref{thm:prior-from-diag-reg} using a combinatorial identity given and proven in \Cref{lem:nice-identity}. 

\begin{proof} (of \Cref{thm:prior-from-diag-reg}).
 By \Cref{lem:gp-map-equiv,lem:gp-equiv-coeff-func}, it suffices to show that $\Phi \Lambda^{-1} \Phi^T =K$. We apply \Cref{lem:nice-identity} and compute  \begin{align*}
    (\Phi \Lambda^{-1} \Phi^T)_{x,y}   &= \sum_{\substack{S,s:\\ x[S]=s,\, y[S]=s}}\Lambda^{-1}_{(S,s),(S,s)} = \sum_{j=0}^{\ell} \sum_{\substack{S,s:\\ x[S]=s,\, y[S]=s\\|S|=j}}\frac{1}{a_j} = \sum_{j=0}^{\ell-d} {\ell-d(x,y) \choose j} \frac{1}{a_j}\\
    &= \sum_{j=0}^{\ell-d} \brac{\sum_{k=0}^j { \ell -k \choose j-k}  \mathcal{K}_k(d(x,y))} \frac{1}{\alpha^j a_j}\\
    &=  \sum_{k=0}^\ell \brac{\sum_{j=k}^\ell \frac{1}{\alpha^j a_j}  {\ell-k \choose j-k}}\mathcal{K}_k(d(x,y)).
  \end{align*}
\end{proof}

\section{A kernel trick for inferring the posterior distribution of transformations of functions}\label{sec:kernel-trick-gen}

For many applications of interest, $\alpha$ and $\ell$ are such that writing down the $\alpha^\ell$-dimensional function vector is impractical. Instead, much can be learned about the function by studying interpretable representations of it. Here we highlight three classes of representations and establish a general kernel trick that can be applied to efficiently compute the posterior distribution of arbitrary subsets of coefficients from these representations under a function space Gaussian process prior specified by a product kernel.  

We will consider the following representations of real-valued functions over sequence space, each of which can be obtained by a linear transformation $M$ of the function vector $f$.
\begin{itemize}
    \item \textbf{Gauge-fixed weights} with respect to a $\lambda$-$\pi$ gauge. The function value of a sequence is given by the sum of weights of its subsequences:  
    
    $$f_x=\sum_{(S,s) \in \mcS} w_{(S,s)} \delta_{x[S]=s}.$$ 
    
    As discussed in  \ref{subsec:lam-pi-fam}, the choice of $\lambda$ and $\pi$ guides the interpretation of the weights $w_{(S,s)}$. Restricting the projection matrix  (\Cref{eq:proj}) to columns corresponding to the whole sequence, $(T,t): |T| =\ell$, gives the linear transformation $M$.  We can compute $w_{(S,s)}= M_{(S,s), \ast} f$ where $M_{(S,s),\ast}$ is a row vector whose formula is given in Table 1. Taking $\lambda =\infty$ and $\pi$ uniform yields the zero-sum gauge, which is utilized in the ``reference-free analysis'' (RFA) approach described in \cite{park2024simplicity}. Taking $\lambda = \infty$ and letting $\pi$ be the point mass distribution on a wild-type sequence ($\pi^p_{WT_p} = 1$) yields the wild-type gauge. As discussed in \cite{posfai2024gauge}, in the wild-type gauge $w_{(S,s)}=0$ for all subsequences $s$ that agree with the wild-type in at least one position; thus, we only define coefficients $w_{(S,u)}$ where $u$ does not agree with the wild-type. 
    \item \textbf{Background-averaged epistastic coefficients.} These coefficients, defined in \cite{poelwijk2016context} for binary alphabets and extended to arbitrary alphabet size in \cite{faure2024extension}, are again defined with respect to a wild-type sequence. For each set of positions $S$ and substring $u$ on $S$ that does not contain any wild-type characters, the corresponding background-averaged epistastic coefficient $\ve_{(S,u)}$ represents the epistastic effect of the combination of mutations $u$ on $S$, averaged over all backgrounds outside the subsequence. Following Equation 9 of \cite{faure2024extension}, we have $\ve_{(S,u)}= M_{(S,u),*}f$  where $M_{(S,u),\ast}$ is a row vector whose formula is given in Table 1. In the bi-allelic ($\alpha =2$) case, the background-averaged epistastic coefficients are a rescaling of the Walsh-Hadamard coefficients (see Equation 6 of \cite{weinberger1991fourier}).

    \item \textbf{Fourier coefficients.} In the Fourier basis~\cite{brookes2022sparsity}, for each subset of positions $S$, there are $(\alpha -1)^{|S|}$ coefficients that together describe the $|S|$-way interaction at those positions. 
    The Fourier basis vectors are the eigenvectors of the Hamming graph Laplacian and are a natural way to express the GNK model \cite{brookes2022sparsity}.
    The Fourier coefficients are also defined with respect to a wild-type sequence, and we index the $(\alpha -1)^{|S|}$ coefficients for the subset of positions $S$ by substrings that do not contain a particular reference (i.e.~wild-type) allele, which without loss of generality we denote as allele zero. Following Equation 10 of \cite{brookes2022sparsity}, we have that each coefficient $\beta_{(S,u)}=  M_{(S,u),*}f$  where $M_{(S,u),\ast}$ is a row vector whose formula is given in Table 1. In the bi-allelic ($\alpha =2)$ case, the Fourier coefficients are the Walsh-Hadamard coefficients (see Equation 6 of \cite{weinberger1991fourier}).
     
\end{itemize}

\noindent  For each of these transformations, naively computing each entry of $Mf$ requires computing an $\alpha^\ell$ dimensional dot product. We leverage the factorizable form of the rows of $M$ to establish a kernel trick that allows us to efficiently compute the posterior distribution of $Mf$.

\begin{table}[ht]
\centering
\begin{tabular}{|p{\dimexpr 0.4\textwidth-2\tabcolsep}|p{\dimexpr 0.6\textwidth-2\tabcolsep}|}
\hline
\textbf{Representation} & \textbf{Transformation matrix}  \\
\hline\hline
\textbf{$\lambda$-$\pi$ gauge-fixed weights \cite{posfai2024gauge}} & $ M_{(S,s),x} = \prod_{p \in S} \brac{\delta_{s_p=x_p}- \pi^p_{x_p}\eta} \prod_{p \not \in S} \pi^p_{x_p} \eta$  \\
\hline
Hierarchical $(\lambda= \infty)$ & $ M_{(S,s),x} = \prod_{p \in S} \brac{\delta_{s_p=x_p}- \pi^p_{x_p}} \prod_{p \not \in S} \pi^p_{x_p}$  \\
\hline
Zero-sum $(\lambda= \infty, \pi \text{ uniform})$, same as RFA in \cite{park2024simplicity} & $ M_{(S,s),x} = \prod_{p \in S} \brac{\delta_{s_p=x_p}- \frac{1}{\alpha}} \prod_{p \not \in S} \frac{1}{\alpha}$  \\
\hline
Wild-type $(\lambda = \infty, \pi  \text{ WT})$  & $M_{(S,u),x} = \prod_{p \in S} \brac{\delta_{x_p=u_p}- \delta_{x_p= WT_p}}\prod_{p \not \in S} \delta_{x_p=WT_p} $\\
\hline\hline
\textbf{Background averaged epistastic coefficients \cite{faure2024extension}} & $M_{(S,u),x} =  \prod_{p \in S} \brac{\delta_{x_p = u_p} - \delta_{x_p=WT_p}}\prod_{p \not \in S} \frac{1}{\alpha}$  \\
\hline
Bi-allelic $(\alpha =2)$, same as rescaled Walsh-Hadamard \cite{weinberger1991fourier}& $M_{S,x} = \frac{1}{2}^{\ell-|S|} \prod_{p \in S} \brac{\delta_{x_p \not= WT_p} - \delta_{x_p=WT_p}}$  \\
\hline\hline
\textbf{Fourier coefficients \cite{brookes2022sparsity}} & $M_{(S,u),x} =\frac{1}{\sqrt{\alpha^\ell}} \prod_{p \in S} \brac{ \delta_{x_p=0}- \frac{1}{\sqrt{\alpha}-1} \delta_{x_p \not = 0} + \sqrt{\alpha} \delta_{x_p= u_p}} $ \\
\hline
 Bi-allelic $(\alpha =2)$, same as Walsh-Hadamard \cite{weinberger1991fourier}& $M_{S,x} =\frac{1}{\sqrt{2^\ell}} \prod_{p \in S} \brac{ \delta_{x_p=0}-\delta_{x_p= 1}} $ \\
\hline
\end{tabular}
\caption{\label{table}The linear transformations $M$ that map the function vector $f$ to a specific representation. Throughout $S$ is used to denote a subset of positions, $s$ is used to denote any subsequence on those positions, and $u$ is used to denote any subsequence on those positions that differs from a wild-type or reference allele at each position. We use $u$ instead of $s$ for the background averaged epistatic coefficients because these coefficients are only defined for subsequences that differ from a wild-type at all positions. We again use $u$ instead of $s$ for the wild-type gauge because all coefficients corresponding to sequences that agree with the wild-type are zero. For bi-allelic alphabets, we suppress the $u$ since there is only one option for such a subsequence. For the $\lambda$-$\pi$ gauges, $\eta = \lambda/(1+\lambda)$. }
\end{table}

For $f$ drawn from a Gaussian process posterior, the distribution of $Mf$ is normal with mean and covariance given in \Cref{thm:trans-kernel-trick}a. While deriving analytic formulas for the mean and covariance of $Mf$ is straightforward, computing the mean and covariance from these formulas requires computing entries of $MK$ and $MKM^T$, which when done naively involves taking $\alpha^\ell$ dimensional dot products. We show that we can compute $MK$ and $MKM^T$ much more efficiently when the Gaussian process prior $K$ is a product kernel and each row of $M$ has a factorizable form  (in the sense of \Cref{eq:factor-M} in \Cref{thm:trans-kernel-trick}b below). In particular, \Cref{thm:trans-kernel-trick}b gives an expression for each entry of $MK$ and $MKM^T$ as the product of $\ell$ factors, each of which is a sum of $\alpha$ or $\alpha^2$ values.
 Given the expressions for $MK$ and $MKM^T$, we can compute each element of the representation $Mf$ with only matrix and vector operations with dimensions at most the number of training sequences.

Since our focus here is on gauge spaces, we dedicate the following section to describing the posterior distribution of gauge-fixed weights and applying \Cref{thm:trans-kernel-trick} to compute an explicit formula for the posterior (\Cref{thm:kernel-trick}). It is easy to verify that the matrices $M$ given in \Cref{table} for the background averaged epistastic coefficients and the Fourier coefficents satisfy the row-factorizable condition (\Cref{eq:factor-M}), and thus \Cref{thm:trans-kernel-trick}b can be applied to efficiently compute the posterior distributions of these representations as well. The value $m^{i,p}_{c}$ is the factor corresponding to position $p$ in the expression in \Cref{table}; in cases where $p \not \in S$ does not appear in the expression, take $m^{i,p}_{c}=1$ for all characters $c$.

\begin{theorem}\label{thm:trans-kernel-trick}
   Let  $f$ be drawn from the posterior of the a function space Gaussian process with covariance $K$ and noise variance $\sigma_n^2$ (see \Cref{defn:fmap} or 1 of \Cref{subsec:equiv}), and let $M$ be a linear transformation of $f$. 
   \begin{enumerate}[(a)]
       \item    The distribution of the random variable $Mf$ is $N(\bar{\theta}, R)$ where $\bar{\theta} = M  K_{*,X} Q y,$ $R =MK M^T - (MK_{*,X})Q(MK_{*,X})^T,\text{ and }  Q=\brac{K_{X,X} + \sigma_n^2 I}^{-1}.$
       \item Let $K$ be a product kernel
       
$$K_{x,y} = \prod_{p=1}^\ell a_{x_p,y_p}^p$$ where $a_{c,c'}^p = a_{c',c}^p$ for all pairs of characters $c$ and $c'$. Suppose each row of $M$ can be factorized by position, \begin{equation}\label{eq:factor-M}
    M_{i,x} = \prod_{p=1}^\ell m^{i,p}_{x_p}.\end{equation}
Then 

$$(MK)_{i,y}=\prod_{p=1}^\ell \brac{ \sum_{c\in \mathcal{A}} m^{i,p}_{c } a^p_{c,y_p}}$$
and

$$(MKM^T)_{i,j} = \prod_{p=1}^\ell \brac{ \sum_{c,c'\in \mathcal{A}} m^{i,p}_{c }  m^{j,p}_{c'}a^p_{c,c'}}.$$

   \end{enumerate}
\end{theorem}

\begin{proof}
    (a) Recall the posterior distribution of the function space Gaussian process, $f \sim N(f^{MAP},C_f)$, where

$$f^{MAP} = K_{*,X} Q y, \quad C_f = K_{*,*} - K_{*,X} Q K_{X,*}, \quad \text{ and } Q=\brac{K_{X,X} + \sigma_n^2 I}^{-1}$$
\Cref{claim:transform-normal} implies $\bar{\theta} = MK_{*,X} Q y$ and 
\begin{align}
R
    = MC_fM^T 
    = MKM^T - (MK_{*,X})Q(MK_{*,X})^T.
\end{align}

(b) Let $\mathcal{S}$ denote the set of sequences. Note that
\begin{align*}
    (MK)_{i,y} = \sum_{x \in \mathcal{S}} \brac{\prod_{p=1}^\ell m^{i,p}_{x_p} a^p_{x_p,y_p}}
    =\prod_{p=1}^\ell \brac{ \sum_{c \in \mathcal{A}} m^{i,p}_c a^p_{c,y_p}},
\end{align*}
as each term in the expansion of the rightmost expression corresponds to a sequence in $\mathcal{S}$.
Next observe
\begin{align*}
(MKM^T)_{i,j} &= \sum_{y \in \mathcal{S}} \brac{\prod_{p=1}^\ell \brac{ \sum_{c \in \mathcal{A}} m^{i,p}_c a^p_{c,y_p}} m^{j,p}_{y_p}}\\
&= \prod_{p=1}^\ell \sum_{c' \in \mathcal{A}} m_{c'}^{j,p} \brac{ \sum_{c \in \mathcal{A}} m^{i,p}_c a^p_{c,c'}}\\
&=\prod_{p=1}^\ell \brac{ \sum_{c,c'\in \mathcal{A}} m^{i,p}_{c }  m^{j,p}_{c'}a^p_{c,c'}}.
\end{align*}
\end{proof}

 \subsection{Kernel trick for inferring the posterior distribution over weights in a $\lambda$-$\pi$ gauge}\label{sec:kernel-trick}

 The results in \Cref{sec:build-it} describe how to perform regularized regression over weight space in a way that induces a chosen function space prior and yields an optimum in a particular gauge. For applications where $\alpha$ and $\ell$ are sufficiently large, a more practical approach is to infer a subset of the $(\alpha +1)^\ell$ gauge-fixed weights corresponding to a curated set of subsequences. 
In this section, we demonstrate how to do so within our framework of function space Gaussian processes using the kernel trick and describe the relationship between gauge-fixing posterior distributions from weight space Bayesian regression and function space Gaussian process regression.

We can map draws from the posterior distribution $f$ to a linear gauge space $\Theta$ by  $\bar{P} f$, where $\bar{P}$ denotes the projection matrix into gauge $\Theta$ (see e.g. \Cref{eq:proj} restricted to columns corresponding to the whole sequence, $(T,t): |T| =\ell$).  Note $\bar{P}$ is $(\alpha +1)^\ell \times \alpha^\ell$ matrix and the coefficients of $f$ in the $\Theta$ gauge are given by $\theta=\bar{P}f$. Note that the distribution of $\theta$ is fundamentally different than the posterior distribution of the \textit{Bayesian regression in weight space} described in \Cref{subsec:equiv}. The former is a distribution over the linear subspace $\Theta$, whereas the latter has support over all of $\R^{(\alpha+1)^\ell}$ and only the MAP estimate $w^{MAP}$ is guaranteed to be in $\Theta$. We can likewise map draws from the posterior distribution of Bayesian regression in weight space to $\Theta$ by applying the associated projection matrix $P$.

\begin{definition}\label{defn:fix-post} Posterior distributions of gauge-fixed weights:
\begin{itemize}
    \item (Function space Gaussian processes). Let  $f$ be drawn from the posterior of the a function space Gaussian process (see \Cref{defn:fmap} or 1 of \Cref{subsec:equiv}). The posterior distribution of gauge-fixed weights is the distribution of the random variable $\bar{P}f$. 
    \item (Weight-space Bayesian regression.) Let  $w$ be drawn from the weight-space Bayesian regression posterior  (see 3 of \Cref{subsec:equiv}). The posterior distribution of gauge-fixed weights is the distribution of the random variable $Pw$. 
\end{itemize}
\end{definition}

The posterior distributions of gauge-fixed weights are normal with singular covariance matrices, ensuring that all draws lie in the gauge-fixed space (see \Cref{eq:gf-func,eq:fg-ws} for the mean and covariance of these distributions). 
 When $K = \Phi W \Phi^T$, i.e. when the function space Gaussian process and the weight space Bayesian regression have corresponding posterior distributions $f$ and $\Phi w$,
 then it is also the case that their posterior distributions over gauge-fixed weights are the same. 

\begin{remark}
   If $K = \Phi W\Phi^T$, then the posterior distributions of gauge-fixed weights corresponding to the function space Gaussian process with covariance $K$ and noise  variance $\sigma_n^2$ and the Bayesian weight space prior with covariance $W$ and noise variance $\sigma_n^2$ are identical.  Note $ P = \bar{P} \Phi$, and so $Pw = \bar{P} \brac{\Phi w}$. By \Cref{lem:gp-equiv-coeff-func}, the posteriors $f$ and $\Phi w$ have the same distribution, and therefore so do $\bar{P}f$ and $\bar{P} \Phi w = P w$. 
\end{remark}

 For a given $K$, there are many possible matrices $W$ for which $K = \Phi W \Phi^T$. Different choices of $W$ will yield different posterior distributions for $w$, with the MAP estimates in different gauges. However, the posterior distribution of gauge-fixed weights $P w$ is the same across all such $W$, further illustrating that the choice of gauge for $W$ does not determine the prior on the learned function.

In the case of $\lambda$-$\pi$ gauges and product kernels it is incredibly efficient to compute the distribution of gauge-fixed weights for any fixed set of subsequences. The following Corollary of \Cref{thm:trans-kernel-trick} gives an explicit formula for the distribution. See \Cref{sec:kernel-cor-comp} for the details of the computation. 

\begin{corollary}\label{thm:kernel-trick}
Let $K$ be a product kernel

$$K_{x,y} = \prod_{p \in P} a_{x_p,y_p}^p$$ where $a_{c,c'}^p = a_{c',c}^p$ for all pairs of characters $c$ and $c'$, and let $\Theta$ be a $\lambda$-$\pi$ gauge. Then the posterior distribution over gauge-fixed weights is given by  $\theta \sim N(\bar{\theta}, R)$ where

$$\bar{\theta}_{(S,s)}=\brac{z^{(S,s)}}^T\brac{K_{X,X} + \sigma_n^2 I}^{-1} y$$ 
and 

$$R_{(S,s),(T,t)} = \prod_{p \in S \cap T} \brac{\bar{\zeta}^p- \zeta^p_{s_p} - \zeta^p_{t_p} + a^p_{s_p,t_p}   }\prod_{p \in S \setminus T} \brac{ \zeta^p_{s_p} - \bar{\zeta}^p }\prod_{p \in T\setminus S} \brac{ \zeta^p_{t_p} - \bar{\zeta}^p } \prod_{p \not \in S \cup T} \bar{\zeta}^p$$
$$-\brac{z^{(S,s)} }^T   \brac{K_{X,X} + \sigma_n^2 I}^{-1} z^{(T,t)},$$
where 

$$\zeta^p_{c} = \eta\sum_{c'} \pi^p_{c'} a_{c,c'}^p, \quad \quad  \bar{\zeta}^p= \eta^2\sum_{c,c'}\pi^p_c\pi^p_{c'}a^p_{c,c'}$$
with $\eta=\lambda/(1+\lambda)$, and $z^{(S,s)}$ is a $|X|$ dimensional vector given by 

$$z^{(S,s)}_x= \prod_{p \in S} \brac{a_{x_p,s_p}^p - \zeta^p_{x_p}}\prod_{p \not \in S}\zeta^p_{x_p}.$$
\end{corollary}

\section{Discussion}

The methods used to represent and infer functions over sequence space can have a substantive impact on interpretation and prediction accuracy \cite{park2024simplicity,dupic2024protein,park2024analysis}. 
Our framework can be used as a guide for computing the implicit function space prior induced by different choices of representation of the function (e.g. gauge-fixed weights \cite{posfai2024gauge} or a basis such as \cite{stadler2000population,brookes2022sparsity,faure2024extension}) combined with a choice of an $L_2$ regularization matrix. Because each combination of representation type and regularizer imposes implicit assumptions, computing the induced function space prior can help practitioners quantify these assumptions, guide their choice of representation and regularizer, and inform their downstream interpretation. Our work further clarifies that although historically $L_2$ regularization on parameter space has been used to both fix the gauge and provide regularization for the estimated function~\cite{posfai2024gauge}, in fact the choice of gauge is a matter of how we choose to represent the learned function and is in principle independent of the form of regularization or Gaussian process prior we impose on function space.

Our main results linking the regularization matrix to the induced function space prior are tailored to our notion of weight space defined by an overcomplete basis of indicator functions. 
To illustrate the importance of the form of the representation, in \Cref{sec:diag-two} we demonstrate that the same regularization matrices behave differently when different bases define the weight space. In particular, we apply our framework to two alternate bases for the bi-allelic case ($\alpha=2$) and show that whereas diagonal regularizers applied when using the Walsh-Hadamard basis induce homoskedastic function space priors, diagonal regularizers applied when using the wild-type basis induce heteroskedastic function space priors. Further investigation is necessary to characterize the assumptions imposed by forms of regularization other than $L_2$, e.g.~$L_1$ regularization  as it is also commonly employed in practice \cite{poelwijk2019learning,brookes2022sparsity,park2024simplicity}.

Our work also provides a theoretically-grounded method for estimating representations of real-valued functions over sequence space and quantifying uncertainty for these estimates. Given a function space prior and a set of training points, our kernel trick can be applied to efficiently compute the posterior distribution for a general class of linear transformations of functions over sequence space. This class includes gauge-fixed weights~\cite{park2024simplicity,posfai2024gauge}, coefficients of the the function when written in the Fourier basis~\cite{brookes2022sparsity}, and background-averaged epistastic coefficients \cite{faure2024extension}, providing efficient estimates and uncertainty bounds for these quantities. Importantly, our kernel trick allows the computation of these estimates without requiring us to explicitly reconstruct the full $\alpha^\ell$-dimensional function, which opens the possibility of computing these quantities on-the-fly in order to quantify higher-order genetic interactions in much longer sequences than have been investigated to date. Moreover, for the case of gauge-fixed weights that is the focus of our contribution here, these individually computed weights maintain the interpretability properties of the chosen gauge (e.g.~the relationship to averages over regions of sequence space~\cite{posfai2024gauge}) and result in a posterior whose support is limited to the $\alpha^\ell$-dimensional gauge space, without ever calculating the full $(\alpha+1)^\ell$ vector of weights.

\section*{Acknowledgements}
This work was supported by a Burroughs-Wellcome Careers at the Scientific Interface (CASI) award (SP), NIH grant R35 GM133613 (CMG, DMM), NIH grant R35 GM133777 (JBK), NIH grant R01 HG011787 (JBK, DMM), and NIH grant R35 GM154908 (JZ), as well as additional funding from the Simons Center for Quantitative Biology at CSHL (DMM, JBK) and the College of Liberal Arts and Sciences at the University of Florida (JZ).
\newpage

\bibliography{refs}
\bibliographystyle{plain}

\appendix

\newpage
\section{Appendix}
\subsection{Useful  lemmas}\label{sec:useful}



\noindent The following is a straightforward computation, included here for completeness. 
\begin{claim}\label{claim:transform-normal}
    Let $x \sim N(\mu, C)$. Then $Px \sim N(P\mu, PCP^T)$.
\end{claim}
\begin{proof} 
Since $Px$ is a linear transformation of a Gaussian random variable, it is also a Gaussian random variable. Linearity implies 
\begin{align*}
Cov(Px) =  \E{(Px-P\mu)(Px-P\mu)^T}= \E{P(x-\mu)(x-\mu)^TP^T} = PCP^T
\end{align*}
\end{proof}

\begin{lemma}\label{lem:gauge-claim-25}(Claim 25 of \cite{posfai2024gauge}.) Let $V_1$ and $V_2$ be complementary subspaces of a vector space $V$. Let $P_1$ be the projection into  $V_1 $ along $V_2$, and $P_2$ be the projection into $V_2$ along $V_1$. Let $\Lambda$ be a symmetric positive-definite matrix acting on $V$. Then the following are equivalent:
\begin{enumerate}
    \item $V_1$ and $V_2$ are $\Lambda$-orthogonal, i.e. $v_1^T \Lambda v_2=0$ for all $v_1 \in V_1$ and $v_2 \in V_2$.
    \item For any fixed $v_1 \in V_1$, $\argmin_{v_2 \in V_2} (v_1+v_2)^T \Lambda (v_1 + v_2) = 0$.
    \item $\Lambda = P_1^T \Lambda P_1 + P_2^T \Lambda P_2$.
\end{enumerate}
\end{lemma}

Next we prove \Cref{lem:form-of-Lambda}, which gives another equivalent condition for $\Lambda$ orthogonality.

\begin{proof}(of \Cref{lem:form-of-Lambda}).
Assume $\Lambda$ orthogonalizes $V_1$ and $V_2$. Since $\Lambda$ is positive-definite, we can write $\Lambda = Z^T Z$ where $Z$ is invertible. Moreover, since $\Lambda$ is orthogonalizing \Cref{lem:gauge-claim-25} implies that 

$$\Lambda = P_1^T\Lambda P_1 + P_2^T \Lambda P_2 = (ZP_1)^TZP_1 + (ZP_2)^TZP_2,$$ where $P_1$ and $P_2$ are the projection matrices into $V_1$ and $V_2$ along $V_2$ and $V_1$ respectively. It remains to show that the null space of $ZP_1$ is  $V_2$ and the null space of $ZP_2$ is $V_1$. Indeed, note that since $Z$ is invertible, if $ZP_1x =0$, then $P_1x=0$, meaning $x$ is in the null space of $P_1$, which is equal to $V_2$. The argument that $V_1$ is the null space of $ZP_2$ is analogous. 

Next assume that $\Lambda = A^TA + B^TB$ where $\nullspace(A) = V_2$ and $\nullspace(B)= V_1$. First we show that $\Lambda$ is positive-definite. Let $v = v_1 +v_2$ where $v_1 \in V_1$ and $v_2 \in V_2$. Note 

$$v^T \Lambda v= v_1^TA^TAv_1 + v_2^TB^TBv_2 = \|Av_1\|_2^2 + \|Bv_2\|_2^2$$
is zero if and only if $v_1 =0$ and $v_2 =0$. 
Next we apply Condition 1 of \Cref{lem:gauge-claim-25} to establish  that $\Lambda$ orthogonalizes $V_1$ and $V_2$. 
Let $v_1 \in V_1$ and $v_2 \in V_2$.  Observe

$$v_1^T(A^TA+B^TB) v_2 = v_1^TA^T(Av_2) +(v_1^TB^T)Bv_2 =0.$$ 
\end{proof}

\begin{claim}\label{claim:phi-delta-phi} Let $\Delta$ be positive-definite. Then  $\Phi^T \Delta \Phi = A^TA$ for some matrix $A$ with nullspace equal to the space of gauge freedoms $G$. 
\end{claim}

\begin{proof}
Since $\Delta$ is positive-definite,  $\Delta = Z^TZ$ for some invertible matrix $Z$, and we can write  $\Phi^T \Delta \Phi  = (Z\Phi)^TZ\Phi$. Note since $Z$ is invertible, $Z\Phi v =0$ implies $\Phi v =0$, so the nullspace of $Z\Phi$ is equal to the nullspace of $\Phi$, which is $G$.    
\end{proof}

\begin{lemma}\label{lem:vc-inverse}
  If $K$ has the form  $K_{x,y} = \sum_{k=0}^\ell \lambda_k \mathcal{K}_k(d(x,y))$, then 

$$K^{-1}_{x,y} = \sum_{k=0}^\ell \lambda_k^{-1} \mathcal{K}_k(d(x,y)).$$
\end{lemma}

\begin{proof}
    This result follows from the fact that $K_{x,y} = \sum_{k=0}^\ell \lambda_k \mathcal{K}_k(d(x,y))$ provides an eigendecomposition of the matrix $K$, see \cite{zhou2022higher}. 
\end{proof}

The following identity is a generalization of equation (22) in  \cite{neidhart2013exact}.

\begin{lemma}\label{lem:nice-identity} Let $j \in [\ell-d]$. Then

$$\alpha^{-j}\sum_{k=0}^j { \ell -k \choose j-k}  \mathcal{K}_k(d; \ell, \alpha) =  { \ell -d \choose j}.$$
\end{lemma}

\begin{proof}
   We will show the equivalent statement
\begin{equation}\label{eq:both-coeffs}
    \sum_{0 \leq i\leq k \leq j} { \ell -k \choose j-k} {d \choose i} {\ell-d \choose k-i}\alpha^j (\alpha-1)^{k-i} (-1)^{i}= {\ell-d\choose j} \alpha^{2j}
\end{equation}
by showing that each side of the \Cref{eq:both-coeffs} is equal to the coefficient of $z^j$ in the polynomial $(\alpha^2z+1)^{\ell-d}$. To see the righthand side, imagine expanding $(\alpha^2z+1)^{\ell-d}$ into $2^{\ell-d}$ terms. Each degree $j$ term has coefficient $\alpha^{2j}$ and there are ${\ell-d \choose j}$ such terms. To see the lefthand side, note that

$$(\alpha^2z+1)^{\ell-d}=(\alpha z + \alpha(\alpha-1)z + 1)^{\ell-d} (\alpha z-\alpha z+1)^d.$$ 
Imagine expanding this polynomial into $3^{\ell}$ terms. We consider the possible coefficients for a degree $j$ term in this expansion and compute how many such terms yield this coefficient. Imagine a degree $j$ term that is the product of $j-k$ copies of $\alpha z$, $k-i$ copies of $\alpha(\alpha-1)z$ and $i$ copies of $(-\alpha z)$. 
There are 

$${\ell-k \choose j-k} {\ell-d \choose k-i} {d \choose i}$$ such terms, each with a coefficient of 

$$ (\alpha )^{j-k} (\alpha(\alpha-1))^{k-i} (-\alpha )^i = \alpha^j(\alpha-1)^{k-i}(-1)^i.$$
Thus, the lefthand side of \Cref{eq:both-coeffs} also counts the coefficient of the degree $j$ term in the polynomial $(\alpha^2z +1)^{\ell-d}.$
\end{proof}

\subsection{Computations that help establish the marginalization property }\label{sec:marg-help-results}

\begin{claim}\label{claim:p-not-in-S}
    Let $b$ and $\mathcal{T}_F$ be defined with respect to a fixed subsequence $(S,s)$ as described in the proof of \Cref{lem:marg-property}. Let $w$ satisfy the $\lambda$-$\pi$ marginalization property.  Then for $p \in S^c \cap F^c$,
    
    $$\sum_{(T,t) \in \mathcal{T}_F} b(T,t) w_{(T,t)} =\sum_{(T,t) \in \mathcal{T}_{F\cup \{p\}}} \frac{b(T,t) w_{(T,t)}}{\eta}.$$
\end{claim}

\begin{proof}
Fix $p \in S^c \cap F^c$. Note that 
$$b(U\cup \{p\}, u^{+c}) = \pi^p_c b(U,u).$$
We compute
\begin{align*}
    \sum_{(T,t) \in \mathcal{T}_F } b(T,t) w_{(T,t)} &= \sum_{(U,u)\in \mathcal{T}_F: p \not \in U}\brac{ b(U,u) w_{(U,u)} + \sum_{c \in \mathcal{A}} b(U\cup \{p\}, u^{+c})w_{(U \cup \{p\}, u^{+c})}}\\
    &=\sum_{(U,u)\in \mathcal{T}_F: p \not \in U}\brac{ b(U,u) w_{(U,u)}  +\sum_{c \in \mathcal{A}}  \pi^p_{c} b(U,u) w_{(U \cup \{p\}, u^{+c})}}\\
    &= \sum_{(U,u)\in \mathcal{T}_F: p \not \in U}   \brac{ b(U,u) w_{(U,u)}  +  \bfrac{1-\eta}{\eta}  b(U,u) w_{(U,u)}}\\
    &= \sum_{(T,t) \in \mathcal{T}_{F\cup \{p\}}} \frac{b(T,t) w_{(T,t)}}{\eta},
\end{align*}
where we used the assumption that $w$ satisfies \Cref{eq:marginalization} to establish the third equality.
\end{proof}    

\begin{claim}\label{claim:p-in-S}
Let $b$ and $\mathcal{T}_F$ be defined with respect to a fixed subsequence $(S,s)$ as described in the proof of \Cref{lem:marg-property}. Let $w$ satisfy the $\lambda$-$\pi$ marginalization property. Then for $p \in S \cap F^c$,
   
    $$\sum_{(T,t) \in \mathcal{T}_F} b(T,t) w_{(T,t)} =\sum_{(T,t) \in \mathcal{T}_{F\cup \{p\}}} \frac{b(T,t) w_{(T,t)}}{1- \pi^p_{s_p}\eta}.$$
    
\end{claim}

\begin{proof}
Fix $p \in S \cap F^c$. Note that 
$$b(U\cup \{p\}, u^{+s_p}) =  \frac{\brac{1-\pi^p_{s_p}\eta } b(U,u) }{1-\eta}=
 \frac{-\pi^p_{s_p} \eta b(U,u) }{1-\eta} + \frac{b(U \cup \{p\}, u^{+s_p}) }{1-\pi^p_{s_p} \eta}, $$
and for $c \not = s_p$
$$b(U\cup \{p\}, u^{+c}) =    \frac{-\pi^p_{c} \eta b(U,u) }{1-\eta}.$$

We compute
\begin{align*}
    \sum_{(T,t) \in \mathcal{T}_F } &b(T,t) w_{(T,t)} = \sum_{(U,u)\in \mathcal{T}_F: p \not \in U}\brac{ b(U,u) w_{(U,u)} + \sum_{c \in \mathcal{A}} b(U\cup \{p\}, u^{+c})w_{(U \cup \{p\}, u^{+c})}}\\
    &=\sum_{(U,u)\in \mathcal{T}_F: p \not \in U}\brac{ b(U,u) w_{(U,u)} + \frac{b(U \cup \{p\}, u^{+s_p}) }{1-\pi^p_{s_p} \eta} w_{(U \cup \{p\}, u^{+s_p})} + \sum_{c \in \mathcal{A}}  \frac{-\pi^p_{c} \eta b(U,u) }{1-\eta}w_{(U \cup \{p\}, u^{+c})}}\\
    &= \sum_{(U,u)\in \mathcal{T}_F: p \not \in U} \frac{b(U \cup \{p\}, u^{+s_p}) }{1-\pi^p_{s_p} \eta} w_{(U \cup \{p\}, u^{+s_p})} \\
    &= \sum_{(T,t) \in \mathcal{T}_{F\cup \{p\}}} \frac{b(T,t) w_{(T,t)}}{1- \pi^p_{s_p}\eta},
\end{align*}
where we used the assumption that $w$ satisfies \Cref{eq:marginalization} to establish the third equality.
\end{proof}

\subsection{Building regularizers for geometric decay, connectedness, and Jenga kernels}

In this section, we discuss three subclasses of product kernels and compute $\Phi^TK^{-1}\Phi$ for each class. 
\subsubsection{Definitions of geometric decay, connectedness, and Jenga kernels}\label{sec:defn-jenga}
We will consider the following three subclasses of product kernels. The most simple subclass are geometric decay kernels, where the covariance decays exponentially with Hamming distance. These isotropic kernels are also a special case of VC kernels corresponding to hyperparameters $\lambda_k$ that decay exponentially with the order $k$ \cite{zhou2022higher, neidhart2013exact}.

\begin{definition}
    A geometric decay kernel has the following form: 
    $$K_{x,y}= \beta^{d(x,y)}$$ where $\beta\in (0,1)$.
\end{definition}

Next we consider connectedness kernels, which are a generalization of the geometric decay kernel that can express that making changes at different positions results in different effects on predictability. The covariation between a pair is the product of site specific factors $z^p$ for all positions $p$ where they differ. In the bi-allelic case ($\alpha = 2)$ when $z^p \in (0,1)$ this prior on function space is equivalent to the connectedness model proposed by \cite{reddy2021global}.
\begin{definition}
    A connectedness kernel has the form 
    
    $$K_{x,y} = \prod_{p: x_p \not =y_p} z^p$$
    where each factor satisfies $\frac{-1}{\alpha-1}< z^p <1 $. 
\end{definition}
\noindent The constraints ensure that $K$ is positive-definite, see \cite{zhou2024tbd}.

Finally, we consider Jenga kernels, a generalization of connectedness kernels that allows different allele-position (character-position) combinations to affect predictability differently. We assign each character-position a factor $z^p_c$; the covariance between a pair of sequences is the product of the factors over the positions where the sequences differ.
\begin{definition}
A Jenga kernel has the form

$$K_{x,y} = \prod_{p: x_p \not = y_p} s_p  z^p_{x_p} z^p_{y_p}$$ where at each position $p$ either
\begin{enumerate}
    \item 
$s_p = 1$ and $z^p_c  \in (0,1)$ for each $c \in \mathcal{A}$, or 
\item $s_p = -1$ and $\sum_{c \in \mathcal{A}} \frac{\brac{z^p_c}^2}{1+ \brac{z^p_c}^2} \leq 1$.
\end{enumerate}
\end{definition}
\noindent The two conditions above correspond to  $z^p>0$ and $z^p \leq 0$ in the connectedness kernel, respectively, and the constraints ensure that $K$ is positive-definite, see \cite{zhou2024tbd}.

\subsubsection{\Cref{thm:phi-kinv-phi-product} applied to geometric decay, connectedness, and Jenga kernels}\label{sec:jenga-cors}
The following corollaries of \Cref{thm:phi-kinv-phi-product} give the form of $\Phi^T K^{-1} \Phi$ for geometric decay, connectedness, and Jenga kernels.
\begin{corollary}
\label{thm:phi-kinv-phi-jenga}
    Let $K$ be a Jenga kernel, $K_{x,y} = \prod_{p: x_p \not = y_p}s^p z^p_{x_p} z^p_{y_p}$,  then $(\Phi^T K^{-1} \Phi)_{(S,s),(T,t)} $ is as given in \Cref{thm:phi-kinv-phi-product}
with 

$$b^p_{c,c'} = \frac{\delta_{c=c'}}{1-s^p\brac{z^p_c}^2} + \frac{s^p\gamma^p z^p_c z^p_{c'}}{\brac{1-s^p\brac{z^p_c}^2}\brac{1-s^p\brac{z^p_{c'}}^2}}, \quad \quad \gamma^p = \frac{-1}{1+ s^p\sum_{c \in \mathcal{A}} \frac{\brac{z^p_c}^2}{1-s^p\brac{z^p_c}^2}}
$$
\end{corollary}

\begin{corollary}\label{cor:phi-kinv-phi-connectedness}
        Let $K$ be a connectedness kernel $K_{x,y} = \prod_{p:x_p \not = y_p} z^p$.  Then
   
    $$(\Phi^T K^{-1} \Phi)_{(S,s),(T,t)} = \prod_{p \in [\ell]} \frac{1}{1+(\alpha-1) z^p} \prod_{\substack{p \in S\cap T\\x_p = y_p}} \frac{1+ (\alpha-2)z^p}{1-z^p}\prod_{\substack{p \in S\cap T\\x_p \not= y_p}} \frac{-z^p}{1-z^p} \prod_{p \not \in S \cup T} \alpha.$$
\end{corollary}

 \begin{corollary}\label{cor:phi-kinv-phi-geometric}
    Suppose $K$ is a geometric decay kernel, $K_{x,y}= \beta^{d(x,y)}$ with $\beta\in (0,1)$. Then 
    
    $$ (\Phi^T K^{-1} \Phi)_{(S,s),(T,t)} =  \frac{\alpha^{\ell-|S\cup T|} }{(1+(\alpha-1)\beta)^\ell}  \bfrac{1+(\alpha -2) \beta}{1-\beta}^{|\{p \in S \cap T: s_p=t_p\}|}
\bfrac{-\beta}{1-\beta}^{|\{p \in S \cap T: s_p\not=t_p\}|}.$$ 
 
\end{corollary}

\Cref{thm:phi-kinv-phi-jenga} follows directly from \Cref{thm:phi-kinv-phi-product} and \Cref{lem:inverse-jenga-piece}, which gives the form of $\brac{K^p}^{-1}$ for a Jenga kernel. 
\begin{lemma}\label{lem:inverse-jenga-piece} Let $K$ be a matrix of the form 

$$K_{ij}= \begin{cases}
    1 & i = j\\
    sa_ia_j & i \not = j
\end{cases}$$
where $s \in \{-1,1\}$.
Then 

$$K^{-1}_{ij} =\frac{ \delta_{i=j}}{(1-sa_i^2)}+\frac{s\zeta a_ia_j}{(1-sa_i^2)(1-sa_j^2)} \quad \text{ for } \quad \zeta = \frac{-1}{1 + s\sum_{i} \frac{a_i^2}{(1-sa_i^2)}}.$$
\end{lemma}

\begin{proof}
The result follows directly from applying Sherman-Morrison formula (\Cref{lem:sherman}) with $A$ the diagonal matrix with $A_{ii} = 1-sa_i^2$ and $u=sv$ with $v_i = a_i$. 
\end{proof}

\begin{lemma}\label{lem:sherman} (Sherman-Morrison formula) Let $A$ be an invertible matrix. Then

$$(A+uv^T)^{-1}=A^{-1} - \frac{A^{-1} uv^T A^{-1}}{1+v^TA^{-1}u}.$$
\end{lemma}

To arrive at the expression in \Cref{cor:phi-kinv-phi-connectedness}, we begin with \Cref{thm:phi-kinv-phi-jenga} and plug in $s^p=sgn(z^p)$ and $z^p_c = \sqrt{|z^p|}$ for all $c$. We obtain

$$b^p_{x_p,y_p} = 
\begin{cases}
    \frac{1 + (\alpha-2) z^p}{(1-z^p)(1+(\alpha-1)z^p)} & x_p = y_p\\
    \frac{- z^p}{(1-z^p)(1+(\alpha-1)z^p)} & x_p \not = y_p
\end{cases}$$

$$\sum_{c \in \mathcal{A}} b^p_{c,c'} = \frac{1 + (\alpha-2)z^p - (\alpha-1)z^p}{(1-z^p )(1+(\alpha-1)z^p)} = \frac{1}{ 1+(\alpha-1)z^p}$$

$$\sum_{c,c'} b^p_{c,c'} = \frac{\alpha(1 + (\alpha-2)z^p) - \alpha(\alpha-1)z^p}{(1-z^p )(1+(\alpha-1)z^p)} = \frac{\alpha}{ 1+(\alpha-1)z^p}$$

To arrive at the expression in \Cref{cor:phi-kinv-phi-geometric}, we plug  $z^p =\beta$ for all $p$ in to \Cref{cor:phi-kinv-phi-connectedness}.

\subsection{Detailed computations for posterior distributions of gauge-fixed weights}\label{sec:pf-of-6-7}

\subsubsection{General formulas for the posterior distributions of gauge-fixed weights}
We compute the posterior distribution of gauge-fixed weights for function space Gaussian processes and weight-space Bayesian regression. Taking $M=\bar{P}$ and applying \Cref{thm:trans-kernel-trick}, we obtain that the posterior distribution of gauge-fixed weights corresponding to a function space Gaussian process with covariance $K$ and noise variance $\sigma_n^2$ is $N(\bar{\theta}, R)$ where
 \begin{equation}\label{eq:gf-func}
     \bar{\theta} = \bar{P}  K_{*,X} Q y, \quad  R = \bar{P} C_f \bar{P}^T, \quad 
     C_f = K - K_{*,X} Q K_{X,*}, \text{ and }  Q=\brac{K_{X,X} + \sigma_n^2 I}^{-1}. 
 \end{equation}
To compute the posterior distribution of gauge-fixed weights corresponding to a Bayesian weight space prior  with covariance $W$ and noise variance $\sigma_n^2$, we first recall the posterior distribution, $w \sim N(w^{MAP},C_w)$, where 
$$w^{MAP} = \sigma^{-2} C_w \Phi^T_X y \quad \text{ and } \quad C_w =\brac{\sigma^{-2} \Phi_X^T\Phi_X + W^{-1}}^{-1}.$$ Applying \Cref{claim:transform-normal} we obtain that the posterior distribution of gauge-fixed weights is $N(\bar{\theta}, R)$ where 
\begin{equation}\label{eq:fg-ws}
    \bar{\theta} = P \sigma^{-2} C_w \Phi^T_X y, \quad R= PC_wP^T, \quad \text{ and } \quad C_w =\brac{\sigma^{-2} \Phi_X^T\Phi_X + W^{-1}}^{-1}.
\end{equation}

\subsubsection{Computations to establish \Cref{thm:kernel-trick}}\label{sec:kernel-cor-comp} 

\begin{proof}(of \Cref{thm:kernel-trick}.) 
Recall 

$$P^{\lambda, \pi}_{(S,s),([\ell],t)} = 
\prod_{p \in S}\brac{\delta_{s_p = t_p} -\pi^p_{t_p} \eta}  \prod_{p \not \in S}\pi^p_{t_p} \eta.
$$
We apply \Cref{thm:trans-kernel-trick}b with $M= \bar{P}$ and 

$$m^{(S,s),p}_c =
\begin{cases}
    \delta_{s_p=c} -\pi^p_c\eta &p\in S\\
    \pi^p_c\eta &p\not\in S.
\end{cases}$$
   The result follows directly from the following computations:
    \begin{align*}
        \brac{\bar{P}K}_{(S,s),y}
        &= \prod_{p \in S} \brac{\brac{1- \pi^p_{s_p}\eta} a_{y_p,s_p}^p - \sum_{c \not = s_p}  \pi^p_{c}\eta a_{y_p,c}^p }   \prod_{p \not \in S} \brac{\sum_{c \in \mathcal{A}} \pi^p_{c}\eta a_{y_p,c}^p}\\
        &= \prod_{p \in S} \brac{a_{y_p,s_p}^p - \zeta^p_{y_p}}\prod_{p \not \in S}\zeta^p_{y_p}
    \end{align*}
    and
    \begin{align*}
        \brac{\bar{P}K\bar{P}^T}_{(S,s),(T,t)}
        &= \brac{\prod_{p \in S\cap T} \sum_{c,c' \in \mathcal{A}}\brac{\delta_{s_p=c}- \pi^p_c \eta}\brac{\delta_{t_p=c'}- \pi^p_{c'} \eta}}
       \brac{ \prod_{p \not \in S \cap T} \sum_{c,c' \in \mathcal{A}} \pi^p_c \eta \pi^p_{c'} \eta}\\
       &         \brac{ \prod_{p \in S \setminus T} \sum_{c,c' \in \mathcal{A}} \brac{\delta_{s_p=c}- \pi^p_c \eta} \pi^p_{c'} \eta}
          \brac{\prod_{p \in T \setminus S} \sum_{c,c' \in \mathcal{A}} \pi^p_c \eta \brac{\delta_{t_p=c'}- \pi^p_{c'} \eta}}\\
          &= \prod_{p \in S \cap T} \brac{\bar{\zeta}^p- \zeta^p_{s_p} - \zeta^p_{t_p} + a^p_{s_p,t_p}   }\prod_{p \in S \setminus T} \brac{ \zeta^p_{s_p} - \bar{\zeta}^p }\prod_{p \in T\setminus S} \brac{ \zeta^p_{t_p} - \bar{\zeta}^p } \prod_{p \not \in S \cup T} \bar{\zeta}^p.
    \end{align*}

    \end{proof}

\subsection{Variance component kernels that cannot be induced by order-dependent diagonal regularizers}\label{sec:cannot-achieve}
Every sequence of positive $\lambda_k$ defines a valid variance component (VC) kernel. Here we show that not all VC kernels can be induced by some order-dependent digaongal regularizer. \Cref{thm:prior-from-diag-reg} establishes that priors induced by order-dependent diagonal regularizers have dimension-normalized $k^{th}$ order variance given by 

$$\lambda_k = \sum_{j=k}^\ell \frac{1}{\alpha^j a_j}   {\ell-k \choose j-k}.$$ Note that such $\lambda_k$'s decrease with $k$, meaning that any sequence of non-decreasing $\lambda_k$'s cannot be induced by an order-dependent diagonal regularizer. Moreover, it is not the case that any VC prior defined by decreasing $\lambda_k$ is induced by an order-dependent diagonal regularizer. 
Indeed note that 

$$\lambda_{\ell-1} = \frac{1}{\alpha^{\ell-1} a_{\ell-1}}+\frac{1}{\alpha^\ell a_\ell}$$
and for $k\leq \ell-1$, 

$$\lambda_k \geq \frac{\ell-k}{\alpha^{\ell-1} a_{\ell-1}} + \frac{1}{\alpha^\ell a_\ell} \geq \lambda_{\ell-1} + \frac{\ell-k-1}{\alpha^{\ell-1} a_{\ell-1}}.$$
This restriction that $\lambda_k$ cannot be arbitrarily close to $\lambda_{\ell-1}$ for order-dependent diagonal regularizers implies that not all VC priors with decreasing $\lambda_k$ can be induced by an order-dependent diagonal regularizer.

\subsection{Function space priors induced by diagonal regularizers for alternative bi-allelic  weight spaces}\label{sec:diag-two}

Finally, we describe the function space priors induced by diagonal regularizers for different weight spaces in the bi-allelic case $(\alpha = 2)$. When $\alpha = 2$, there are natural bases to use for regularized regression that are interpretable and not overparameterized \cite{weinberger1991fourier,poelwijk2016context}: the Walsh-Hadamard basis  (WH) and the wild-type basis (WT). Both have one basis element associated with each subset of positions $S \subseteq [\ell]$, but the meaning of the weight of each basis element is interpreted differently.

\paragraph{Walsh-Hadamard basis.} We encode the alleles as $\{-1, +1\}$ and write a function $f$ in terms of $2^\ell$ weights $w_S$ as 

$$ \quad f = \sum_S w_S \prod_{p \in S} x_p = Hw \quad \text{ where } \quad H_{x,S} = \prod_{p \in S} x_p.$$ 

\paragraph{Wild-type basis.} We encode the wild-type allele as $0$ and the mutant allele as $1$. We write a function $f$ in terms of $2^\ell$ weights $w_S$ as 

$$f=\sum_S w_S \prod_{p \in S} \delta_{x_p=1}= Tw \quad \text{ where } \quad T_{x,S} = \prod_{p \in S}\delta_{x_p=1}. $$

We apply the framework established in \Cref{subsec:equiv} to describe how regularized regression in these $\alpha^\ell=2^\ell$-dimensional weight spaces induce function space priors. Here $H$ or $T$ will play the role of $\Phi$.  We show that any diagonal regularizers with the WT basis induces a heterosketdastic prior, whereas any diagonal regularizer with the WH basis induces a homoskedastic prior. Moreover, we describe how a subset of connectedness kernels can be induced with diagonal regularizers with the WH basis.  

\begin{theorem}\label{thm:2-allele}
    Let $\Lambda$ be diagonal regularizer indexed by subsets of positions,

$$\Lambda_{S,S} = \prod_{p \in S} \rho_p$$
where $\rho_p >0$. Let $f^{MAP}(K, \sigma_n^2)$ be the MAP estimate for the Gaussian process $y = f_X +\ve$ where $f \sim N(0,K)$ and $\ve \sim N(0, \sigma_n^2 I)$. 
\begin{enumerate}
    \item When used with the WH basis, the regularizer $\Lambda$ induces the function space prior 
    
    $$K^{(H)}_{x,y}=\brac{ \prod_p \brac{1+ \rho_p^{-1}}} \prod_{p:x_p \not = y_p} \frac{1-\rho_p^{-1}}{1+\rho_p^{-1}},$$ meaning 
    
    $$Hw^{OPT}(\Lambda, \sigma_n^2) = f^{MAP}(K^{(H)}, \sigma_n^2)$$ where 
    
    $$w^{OPT}(\Lambda, \sigma_n^2) = \argmin_{w \in \R^{2^\ell}} \| y-H_Xw\|_2^2 + \sigma_n^2 w^T \Lambda w.$$
    \item When used with the WT basis, the regularizer $\Lambda$ induces the function space prior 
    
    $$K^{(T)}_{x,y}=\prod_{\substack{p \in S:\\ x_p = y_p = 1}} (1+\rho_p^{-1}),$$ meaning 
    
    $$Tw^{OPT}(\Lambda, \sigma_n^2) = f^{MAP}(K^{(T)}, \sigma_n^2)$$ where 
    
    $$w^{OPT}(\Lambda, \sigma_n^2) = \argmin_{w \in \R^{2^\ell}} \| y-T_Xw\|_2^2 + \sigma_n^2 w^T \Lambda w.$$
\end{enumerate}
\end{theorem}

Whereas $K^H$ is a homoskedastic prior, $K^{(T)}$ is heteroskedastic; the variance of the wild-type sequence is one and the variance of the other sequences depends on the extent to which they differ from the wild-type (sequences that are more different from the wild-type tend to have higher variances). We can induce a connectedness prior with $0<z^p<1$ through regularized regression in the WH basis by choosing $\rho_p = (1+z^p)/(1-z_p)$ and rescaling by $1/ \prod_p (1+ \rho_p^{-1})$.

\begin{proof} (of \Cref{thm:2-allele}).
Taking $\Phi$ to be $H$ or $T$, we apply \Cref{lem:gp-map-equiv,lem:gp-equiv-coeff-func} to conclude that the regularizer $\Lambda$ induces the kernels $H\Lambda^{-1} H^T$ and $T\Lambda^{-1} T^T$ on function space for the WH and WT bases respectively. 
    Observe
    \begin{align*}
    (H\Lambda^{-1}H^T)_{x,y}&= \sum_S \Lambda^{-1}_S \prod_{p \in S} x_p y_p\\
    &=\sum_S \brac{ \prod_{p \in S} \rho_p^{-1} } \brac{\prod_{\substack{p \in S\\x_p \not = y_p}}(-1)}\\
    &= \prod_{p: x_p = y_p} (1+\rho_p^{-1})\prod_{p: x_p \not = y_p} (1- \rho_p^{-1})\\
    &= \brac{ \prod_p \brac{1+ \rho_p^{-1}}} \prod_{p:x_p \not = y_p} \frac{1-\rho_p^{-1}}{1+\rho_p^{-1}}.
    \end{align*}

Similarly
    \begin{align*}
    (T\Lambda^{-1}T^T)_{x,y}= \sum_S \Lambda^{-1}_S \prod_{p \in S} \delta_{x_p=1}\delta_{y_p=1}
    =\sum_S \brac{ \prod_{\substack{p \in S:\\ x_p = y_p = 1}} \rho_p^{-1} } 
    = \prod_{\substack{p \in S:\\ x_p = y_p = 1}} (1+\rho_p^{-1}).
    \end{align*}
\end{proof}

\end{document}